\documentclass[11pt]{article}

\usepackage[round]{natbib}

\usepackage{fullpage}

\usepackage{picture,bm}
\usepackage{xcolor}
\usepackage{url}
\usepackage[colorlinks=true,linkcolor=blue,citecolor=black,pagebackref=true]{hyperref}  

\usepackage{dsfont}
\newcommand{\indic}{\mathds{1}}

\usepackage{amsmath,amsthm,amstext,amsfonts,amssymb,mathrsfs}
\usepackage{graphicx,caption,subcaption,algorithm,algorithmic} 

\usepackage{wrapfig,bbm,booktabs}

\usepackage[section]{placeins}

\usepackage{times}

\newtheorem{assumption}{Assumption}

\newtheorem{theorem}{Theorem}

\newtheorem{remark}{Remark}

	\definecolor{airforceblue}{rgb}{0.36, 0.54, 0.66}
	\definecolor{burntorange}{rgb}{0.8, 0.33, 0.0}
	\definecolor{blue}{rgb}{0.0, 0.0, 1.0}

\DeclareMathOperator*{\argmin}{arg\,min}

\newcommand{\redtext}[1]{{\leavevmode\color{red}#1}}

\title{\bf Off-Policy Evaluation Using Information Borrowing and Context-Based Switching}

\author{ 
   Sutanoy Dasgupta \thanks{\small Equal contribution. This work was done primarily when all authors were at Texas A\&M University. This version includes comparisons with more methods (Nadaraya-Watson regression) and updated related work discussions in the Introduction. Email:  \texttt{SD: sutanoy26071991@gmail.com, YN: yniu4@uh.edu, KP: kpb@caltech.edu, DK: dileep.kalathil@tamu.edu, DP: debdeep@stat.tamu.edu, BM: bmallick@stat.tamu.edu}} \\
   {\small Senior Data Scientist}\\
   {\small Intuit, Bengaluru, India}\\
   \and
   Yabo Niu$^*$ \\
    {\small Department of Mathematics}\\
   {\small University of Houston}\\
   \and
   Kishan Panaganti$^*$ \\
   {\small Computing + Mathematical Sciences Department} \\
    {\small California Institute of Technology}\\
   \and
    Dileep Kalathil  \\
   {\small Department of Electrical and Computer Engineering} \\
    {\small Texas A\&M University}\\
     \and
   Debdeep Pati \\
    {\small Department of Statistics}\\
   {\small Texas A\&M University}\\
   \and
    Bani Mallick\\
    {\small Department of Statistics}\\
   {\small Texas A\&M University}  
}

\date{}

\begin{document}
\maketitle

\begin{abstract}
We consider the off-policy evaluation (OPE) problem in contextual bandits, where the goal is to estimate the value of a target policy using the data collected by a logging policy. Most popular approaches to the OPE are variants of the doubly robust (DR) estimator obtained by combining a direct method (DM) estimator and a correction term involving the inverse propensity score (IPS). Existing algorithms primarily focus on strategies to reduce the variance of the DR estimator arising from large IPS. We propose a new approach called the  \underline{D}oubly \underline{R}obust with \underline{I}nformation borrowing and \underline{C}ontext-based switching (DR-IC) estimator that focuses on reducing both bias and variance. The DR-IC estimator replaces the standard DM estimator with a parametric reward model that borrows information from the `closer’ contexts through a correlation structure that depends on the IPS. The DR-IC estimator also adaptively interpolates between this modified DM estimator and a modified DR estimator based on a context-specific switching rule. We give provable guarantees on the performance of the DR-IC estimator. We also demonstrate the superior performance of the DR-IC estimator compared to the state-of-the-art OPE algorithms on a number of benchmark problems.
\end{abstract}

\section{Introduction}
\label{sec:introduction}

Contextual bandits framework finds a wide range of applications such as  recommendations systems \citep{li2011unbiased}, personalized healthcare \citep{zhou2017residual}, advertising \citep{bottou2013counterfactual}, and education \citep{mandel2014offline}.  In contextual bandits, a decision maker (often an algorithm) observes a context and takes an action according to a policy and observes a reward. For a recommendation algorithm,  the context can be the information about the user and the history is  his past online  clicks/purchases, and a reward is obtained when he clicks/buys the item. Importantly,  the algorithm  can only observe the reward for the selected  actions, and not for the others.  The goal  is to maximize the expected reward by choosing the  best policy out of a set of candidate policies.  A natural approach towards that end is to  evaluate the candidate policies by calculating their expected reward, called {policy evaluation}, and then choose the policy with the highest expected reward. We consider the fundamental problem of {off-policy evaluation} (OPE) in contextual bandits, where one uses the data gathered by a past policy, known as the logging policy, to estimate the expected reward of a new policy, known as the target policy. OPE eliminates the cost of performing policy evaluation  via online experiments, and avoids the risks  of exposing subjects to untested policies.

There are three main approaches to address the OPE problem: $(i)$ \textit{Direct Method (DM)}  first gets an estimate of the reward model and uses it to estimate the value of the target policy. DM often has low variance. However, its  bias  can be significant, especially when the true reward model does not belong to the function class used for the reward function estimation. $(ii)$ \textit{Inverse  Propensity Scoring (IPS)} estimator \citep{horvitz1952generalization} uses importance weighting to correct the mismatch between proportions of actions due to the difference between the logging and target polices. While the IPS estimator is unbiased, its variance can be really large due to large importance weights. $(iii)$ \textit{Doubly Robust} (DR) estimator \citep{robins1995semiparametric, bang2005doubly, dudik2011doubly, dudik2014doubly} is a combination of DM and IPS to achieve the low variance of DM and no bias of IPS. While DR is known to be asymptotically optimal  under some assumption \citep{su2020doubly},  its finite-sample variance can still be quite high when importance weights are large.

Many recent works focus on reducing the variance of the DR estimator. \citet{wang2017optimal} introduced a switching approach that switches between using a DM and DR, depending on the importance weights. \citet{su2020doubly} takes a shrinkage approach where they replace the importance weights by smaller weights which optimizes the mean squared error. 
The variance of such estimators is usually caused by mismatches in supports of data-collecting policy and target policy.
\citet{mou2023kernel} assume that the reward function lies within a reproducing kernel Hilbert space to address such mismatches. 
\citet{khan2024off} instead use partial identification methods to address these mismatches.
There are however few works addressing the high bias of the DM method. \citet{su2020doubly} has revealed the importance of choosing an appropriate reward model. Through an extensive study, they have shown that different reward models result in the best estimates in different scenarios. \citet{lichtenberg2023double} use reward clipping approaches for achieving practical less-bias results.

In this work, we propose an innovative approach for  the OPE problem,  which we call the \underline{D}oubly \underline{R}obust with \underline{I}nformation borrowing and \underline{C}ontext-based switching (DR-IC) estimator. The approach is based on two key proposals for improving the performance of the standard DR estimator.

First, we propose a modified DM that uses an intuitive and widely applicable parametric reward model that can significantly reduce the bias. The fundamental idea of the proposed reward model is {\it borrowing information} from `similar' and  `important' contexts through a correlation structure. In particular, this reward model assigns more weight to data points that are `similar' to the test data point through the choice of a kernel. The bandwidth of the kernel is chosen adaptively, depending on the importance weight associated with the test data point. We call this new DM as DM with  \underline{I}nformation \underline{B}orrowing (DM-IB). We theoretically show that the bias of the DM-IB  goes to zero as the number of samples increases. We also empirically demonstrate  the advantages of DM-IB.

Second, we propose a context-based switching scheme  for adpatively choosing between the DM-IB and the DR estimator (which also uses a reward model with information borrowing). Unlike the switching scheme proposed in \cite{wang2017optimal} which does switching based on the importance weights, our approach uses the  Kullback-Leibler (KL) divergence between the logging policy and target policy, given the context, to do the switching. We call this  the DR-IC estimator. Through extensive experiments,  we demonstrate the superior performance of our DR-IC estimator compared to the state-of-the-art methods \citep{wang2017optimal, farajtabarCG18, su2020doubly} on a number of benchmark problems.

\section{Preliminaries}
\label{sec:formulation}

In the contextual bandits problem, the algorithm observes a context $x \in \mathcal{X}$, takes an action $a \in \mathcal{A}$, and gets a scalar reward $r(x, a)$. We assume that the context  space $\mathcal{X} \subset \mathbb{R}^{d}$ is continuous and the action space $\mathcal{A}$  is finite. The contexts   are sampled i.i.d.  according to a distribution $\nu_{C}(\cdot)$. Reward $r(x,a)$ has a distribution conditioned on $(x, a)$ denoted by $\nu_{R}(\cdot|x, a)$. We assume that $\nu_{R}(\cdot|x, a)$ is a Lipschitz function with respect to $x$ for any given $a$, formalizing the intuition that similar context should give similar rewards for the same action.   The algorithm selects actions according a decision rule called policy, which maps the observed context to a distribution over the action space. A policy $\pi$ is denoted as a conditional distribution $\pi(a|x)$ which specifies the probability of selecting action $a$ when the context $x$ is observed.  The value of a policy $\pi$, denoted as $V^{\pi}$, is defined as
\begin{align}
    \label{eq:value-defn}
    V^{\pi} = \mathbb{E}_{x \sim \nu_{C}} \mathbb{E}_{a \sim \pi(\cdot|x)}  \mathbb{E}_{r \sim \nu_{R}(\cdot|x,a)} [r].
\end{align}
In the following, we will also write $\pi(x, a, r)$ to denote the joint distribution over context-action-reward tuples when actions are selected by the policy $\pi$. i.e., $\pi(x, a, r) =  \nu_{C}(x) \pi(a|x) \nu_{R}(r|x, a)$.

In the off-policy evaluation problem, we are given a data set $\mathcal{D} = \{(x_{i}, a_{i}, r_{i})\}^{n}_{i=1}$ that consists of $n$ i.i.d. samples of  context-action-reward tuple generated by a \textit{logging policy} $\mu$, i.e., $(x_{i}, a_{i}, r_{i}) \sim \mu$.  The goal is to estimate the value $V^{\pi}$ of a \textit{target policy} $\pi$ using this offline dataset $\mathcal{D}$.

There are three standard and well known approaches for off-policy evaluation. 

\emph{Direct Method:} In DM, we train a reward model $\widehat{r}$ as
\begin{align}
\label{eq:rhat-estimator-DM}
    \widehat{r}(x, a) = \argmin_{f \in \mathcal{F}} \sum^{n}_{i=1} (f(x_{i}, a_{i}) - r_{i})^{2},
\end{align}
where $\mathcal{F}$ is a suitable function class. The  DM estimator is then given as
\begin{align}
\label{eq:DM-estimator}
    \widehat{V}^{\pi}_{\mathrm{DM}} = \frac{1}{n} \sum^{n}_{i=1} \sum_{a \in \mathcal{A}} \pi(a | x_{i}) \widehat{r}(x_{i}, a).
\end{align}
Though the DM estimator typically has low variance, it often suffers from  large  bias \citep{dudik2011doubly}. 

\emph{Inverse Propensity Scoring Estimator :}  IPS estimator makes use of the \textit{importance weights}, defined as $w(x, a) := \frac{\pi(a|x)}{\mu(a|x)}$, to get an unbiased estimate \citep{horvitz1952generalization}. The IPS estimator is given as
\begin{align}
\label{eq:IPS-estimator}
    \widehat{V}^{\pi}_{\mathrm{IPS}} = \frac{1}{n} \sum^{n}_{i=1} w(x_{i}, a_{i}) r_{i}.
\end{align}
We make the standard assumption that $\mu(a|x) > 0$ whenever  $\pi(a|x) > 0$, ensuring that $w(x,a) < \infty$. Though the IPS estimator is unbiased, when there is a substantial mismatch between the policies $\mu$ and $\pi$, the importance weights will be large and hence the variance will also be large. 

\emph{Doubly Robust Estimator:} DR approach combines DM and IPS estimator as
\begin{align}
    \label{eq:DR-Estimator}
    \widehat{V}^{\pi}_{\mathrm{DR}} = \widehat{V}^{\pi}_{\mathrm{DM}} + \frac{1}{n} \sum^{n}_{i=1} w(x_{i}, a_{i}) ( r_{i} - \widehat{r}(x_{i}, a_{i})),
\end{align}
where $\widehat{r}$ is as given in \eqref{eq:rhat-estimator-DM}. DR estimator preserves the unbiased nature of the IPS and hence is robust to a poor reward estimator $\widehat{r}$.  At the same time, because the IPS part in the DR estimator is using a shifted reward, the variance of the DR estimator is smaller than that of the IPS estimator.  It is known that DR estimator is asymptotically optimal as long the reward estimator is consistent \citep{su2020doubly}.  However,  its finite-sample variance can  be quite high
due to large importance weights when the logging policy and the evaluation policy are different. 

Recently, different approaches to further reduce the variance of the DR approaches have been studied. The closest to our work is \citep{wang2017optimal}, which proposed a switching approach based on the importance weights. The proposed  switch estimator uses the IPS approach when the importance weights are less than a threshold and switches to the DM approach when the importance weights exceeds that threshold. Formally, switch estimator is given as
\begin{align}
    \label{eq:switch-estimator}
  \widehat{V}^{\pi}_{\mathrm{S}} &=  \frac{1}{n} \sum^{n}_{i=1} w(x_{i}, a_{i}) r_{i} \indic\{w(x_{i}, a_{i}) \leq \tau\} 
   + \nonumber
   \\&\frac{1}{n} \sum^{n}_{i=1} \sum_{a \in \mathcal{A}} \pi(a | x_{i}) \widehat{r}(x_{i}, a) \indic\{w(x_{i}, a_{i}) > \tau\},
\end{align}
where $\tau$ is the threshold. The IPS part in the above estimator is can also be replaced by a DR estimator.

\section{Our Approach}
\label{sec:algorithm}

In this section, we introduce our new approach, which we call the \underline{D}oubly \underline{R}obust with \underline{I}nformation borrowing and \underline{C}ontext-based switching (DR-IC) estimator. The approach is based on two key proposals for improving the performance of the standard DR estimator. First, we propose a modified DM that uses a innovative reward model which borrows information from similar and important contexts in order to reduce the bias. We call this as DM with Information Borrowing (DM-IB).  Second, we propose a context-based switching scheme  for adpatively choosing between the DM-IB  and the DR (which also uses the information borrowing  reward model) for reducing the variance. We call this combined approach the DR-IC estimator.  

\subsection{Direct Method with  Information Borrowing Reward Model}

A major drawback of traditional DM estimators is that they do not exploit  the information about the target policy while training a reward model $\widehat{r}$ as given in \eqref{eq:DM-estimator}. So, each data sample $(x_{i}, a_{i}, r_{i})$ is used uniformly in \eqref{eq:DM-estimator}. However, depending on the importance weight, some of the samples are more relevant than others with respect to the off-policy evaluation objective. For example, if the importance weight $w(x_{i}, a_{i})$ is large for sample $i$,  this sample is more important in estimating $V^{\pi}$ than another sample with smaller importance weight. While performing a standard weighted regression with importance weights may address this problem partially, it does not exploit another important fact that ``similar'' contexts should produce ``similar'' rewards under the same action. So, a good reward model should be able to borrow more information from contexts that are ``similar'' than contexts that are different, in order to reduce the bias in estimation.

A standard reward model estimate used in the literature is least squares regression with $\ell_{2}$ regularization and Gaussian errors \citep{dudik2011doubly,wang2017optimal, su2020doubly}. Let $z_i=(x_i^{\top}, a_i)^{\top}$ be the context-action pairs, $i=1,2, \ldots, n$,  $Z = {[z_{1}, \ldots, z_{n}]}^{\top}$ be the matrix of context-action pairs, and $\mathbf{r} = {[r_{1}, \ldots, r_{n}]}^{\top}$ be the vector of rewards obtained from the data $\mathcal{D}$. The reward estimate for any context-action pair $z=(x^\top, a)^\top$ according  least squares regression with $\ell_{2}$ regularization is given by
\begin{align}
\label{eq:ls-estimate}
    \widehat{r}_{\mathrm{ls}}(z) = z^{\top} \widehat{\theta}_{\mathrm{ls}}, ~\text{where}~ \widehat{\theta}_{\text{ls}} = {(Z^T Z + \lambda I)}^{-1}Z^T \mathbf{r}. 
\end{align}

We now consider a \textbf{derived data set} $\widetilde{\mathcal{D}} = \{(\widetilde{x}, \widetilde{a} ) : \widetilde{x} \in \mathcal{D}, \widetilde{a} \in \mathcal{A}\}$. Note that the cardinality of $\mathcal{D}$ is $n$ and  of $\widetilde{\mathcal{D}}$ is $n |\mathcal{A}|$. Let $\widetilde{z} = (\widetilde{x}^{\top}, \widetilde{a})^{\top}$ be the context-action pair  corresponding to the sample  $(\widetilde{x}, \widetilde{a} )$ from the set  $\widetilde{\mathcal{D}}$ and let $\widetilde{Z} = {[\widetilde{z}_{1}, \ldots, \widetilde{z}_{n|\mathcal{A}|}]}^{\top}$. Our goal is to get a reward model that gives a reward estimate for each sample $\widetilde{z}$ using correlation between $\widetilde{\mathbf{r}}$ and $\mathbf{r}$, where $\widetilde{\mathbf{r}}$ is the {\it predicted} rewards given a context-action pair $\widetilde{z}$.  Note that the derived dataset comprises of the original datapoints $\mathcal{D}$ paired with all possible actions $\mathcal{A}$, as we have  explained in Section 3. We emphasize that \textit{we do not need additional data} to implement our algorithm.

Since we don't have an explicit correlation structure available a priori, we  model the covariance matrix between $\widetilde{\mathbf{r}}$ and  $\mathbf{r}$, $\Sigma(\widetilde{\mathbf{r}}, \mathbf{r})$, as
\begin{align} 
\label{covf}
    [\Sigma(\widetilde{\mathbf{r}}, \mathbf{r})]_{ji} =  \frac{1}{h_n\sqrt{\widetilde{w}_jw_i}}K \left( \frac{{\| \widetilde{z}_{j} - z_{i}  \|_2}}{  h_n\sqrt{\widetilde{w}_{j} w_{i}}   }\right),
\end{align}
where $w_{i} = w(x_{i}, a_{i})$, $\widetilde{w}_{j} = w(\widetilde{x}_{j}, \widetilde{a}_{j})$, $j=1, \ldots, n|\mathcal{A}|$, $i = 1, \ldots, n$, and $h_n$ is the bandwidth  and $K(\cdot)$ is an appropriate kernel.

The main motivation for defining such a covariance matrix is to exploit the fact that  similar contexts should give similar rewards under the same action. Thus, the proposed covariance structure borrows more information from similar contexts due to the term $\| \widetilde{x}_{j} - x_{i}\|_{2}$. Also, if $w_{i}$ or $\widetilde{w}_{j}$ are large, these samples are more important for the off-policy evaluation. So, our covariance structure  borrows more information from such samples.  In contrast, a standard least squares regression approach as given in \eqref{eq:ls-estimate}  does not consider the similarity of contexts and borrows information uniformly from all samples, resulting in a higher bias. For our experiments, we take a specific covariance matrix  of the form \eqref{covf}, namely a  Gaussian kernel truncated at the same actions, as given by 
\begin{align} 
\label{eq:gaussian-covf}
  \frac{1}{\sqrt{2\pi}h_n\sqrt{\widetilde{w}_jw_i}} \exp \left(- \frac{{\| \widetilde{x}_{j} - x_{i}  \|^{2}_{2}}}{2 h_n^2\widetilde{w}_{j} w_{i}   }\right)\indic(\widetilde{a}_{j} = a_{i} ).
\end{align}

Let $\Sigma_{\mathbf{r}}$ be the covariance matrix for the original dataset. Assume that the true underlying covariance structure between $\widetilde{\mathbf{r}}$ and  $\mathbf{r}$ is given by \eqref{covf}. Then,  the conditional mean of $\widetilde{\mathbf{r}}$, denoted by $\widehat{\mathbf{r}}_\mathrm{IB}$, given the original rewards $\mathbf{r}$, the context-action pairs $Z$ and $\widetilde{Z}$ and the regression parameter $\widehat{\theta}_{\mathrm{ls}}$ is our reward model (estimate) for the proposed approach:
\begin{align} 
 & \widehat{\mathbf{r}}_{\mathrm{IB}}(\widetilde{Z}) =\widetilde{Z}\widehat{\theta}_{\mathrm{ls}} + \label{eq:r-IB-1} \\\nonumber&\hspace{1.3cm} \Sigma(\widetilde{\mathbf{r}}, \mathbf{r})\Sigma_\mathbf{r}^{-1}\left[\mathrm{diag}\{\Sigma(\widetilde{\mathbf{r}}, \mathbf{r})\Sigma_\mathbf{r}^{-1}\bm{1}_n\}\right]^{-1}(\mathbf{r} - Z\widehat{\theta}_{\mathrm{ls}}),
\end{align}
where $\bm{1}_n$ is an $n$-dimensional vector which all of its entries are 1. The first term in the equation is simply the linear regression estimate if the correlation structure was assumed to be non-existent, as in usual linear regression. The second term is the \textit{information borrowing} term which captures the borrowing of information based on the similarity among the contexts in the original dataset and the `derived' dataset. The conditional expectation of $\widetilde{\mathbf{r}}$ given $\mathbf{r}$ as a projection of $\widetilde{\mathbf{r}}$ is a natural estimator for $\widetilde{\mathbf{r}}$ and is the best estimator among all functions $f(\mathbf{r})$ of $\mathbf{r}$ in the sense that it minimizes the mean squared error (MSE), $\mathbb{E}[\|(\widetilde{\mathbf{r}} - f(\mathbf{r}))\|^2]$. We refer the reader to  \cite{brunk1961best,brunk1963extension} and \cite{vsidak1957relations} for more details. In practice, we do not know $\Sigma_\mathbf{r}$ and hence we replace it with an estimate. The scaling term $\left[\mathrm{diag}\{\Sigma(\widetilde{\mathbf{r}}, \mathbf{r})\Sigma_\mathbf{r}^{-1}\bm{1}_n\}\right]^{-1}$ ensures that the underlying joint covariance structure of $\widetilde{\mathbf{r}}$ and $\mathbf{r}$  is positive definite.

Using the $ \widehat{\mathbf{r}}_{\mathrm{IB}}$ as the reward estimator, we now propose  a new  direct method for off-policy evaluation called DM-IB, given by 
\begin{align}
\label{eq:V-DM-IB}
\widehat{V}^{\pi}_{\mathrm{DM\mbox{-}IB}} = \frac{1}{n}\sum_{i=1}^n\sum_{a\in \mathcal{A}}\pi(a\mid x_i)\,\widehat{r}_{\mathrm{IB}}(a,x_i).
\end{align}
In the Section \ref{sec:analysis}, we will show the superior properties of the proposed estimator $\widehat{V}^{\pi}_{\mathrm{DM\mbox{-}IB}}$ compared to the standard direct method  using the reward estimator as given in \eqref{eq:ls-estimate}. In particular, we will show that the bias of $\widehat{V}^{\pi}_{\mathrm{DM\mbox{-}IB}}$ converges to zero in probability as $n$ increases.

%

\subsection{Doubly Robust Method with Context-Based Switching}

While borrowing information through a correlation structure to get a reward estimator as proposed in \eqref{eq:r-IB-1}  can reduce the bias compared to the standard DM method, the variance can  still be significant compared to  the standard doubly robust approach. On the other hand, simply using the proposed reward estimator in a doubly robust structure alone will not reduce the high variance  of the DR estimator arising due to the difference between the logging policy and evaluation policy (and  reflected in the large importance weights). To overcome this challenge, we propose  a switching approach that adaptively selects between DM-IB and  DR estimator. In particular,  we propose to perform the  switching  based on the context-specific KL divergence between the logging policy and evaluation policy.

The KL divergence $D_\mathrm{KL}$ between two probability mass functions $p_1$ and $p_2$ is given by $
D_\mathrm{KL}(p_1 || p_2) = \sum_x p_1(x) \log \frac{p_1(x)}{p_2(x)}.$
We define the context specific KL divergence $D_\mathrm{KL}(x)$ for a given context $x$, as $
    D_\mathrm{KL}(x) = D_\mathrm{KL}(\pi(\cdot \mid x  )~ ||~ \mu(\cdot \mid x)).$ Our DR-IC estimator $\hat{V}^{\pi}_\mathrm{DR\mbox{-}IC}$ is now given as
\begin{align}
&\frac{1}{n}\sum_{i=1}^n (r_i - \widehat{r}_\mathrm{IB}(x_i,a_i))  \frac{\pi (a_i \mid x_i)}{\mu (a_i \mid x_i)})\,\indic(D_\mathrm{KL}(x_i) <\tau) \nonumber \\
&+ \frac{1}{n}\sum_{i=1}^n \sum_{a \in \mathcal{A}}  \widehat{r}_\mathrm{IB}(x_i, a)\pi( a \mid x_i)\,\indic(D_\mathrm{KL}(x_i)
< \tau) 
\nonumber \\
\label{eq:v-dr-ic}
&+\frac{1}{n}\sum_{i=1}^n \sum_{a \in \mathcal{A}} \widehat{r}_\mathrm{IB}(x_i, a)\pi( a \mid x_i)\,\indic(D_\mathrm{KL}(x_i)
\geq \tau),
\end{align}
where $\tau$ is the threshold parameter that determines the switching.  The third term  in \eqref{eq:v-dr-ic} is exactly our new direct method estimate $\widehat{V}^{\pi}_{\mathrm{DM\mbox{-}IB}}$ given \eqref{eq:V-DM-IB}.  The first two terms in \eqref{eq:v-dr-ic} constitute a DR estimator which uses the information borrowing reward estimate  $\widehat{r}_\mathrm{IB}$ as proposed in  \eqref{eq:r-IB-1}. The DR estimator is used only for contexts with  KL divergence less than the threshold $\tau$, and we switch to the DM method if the KL divergence exceeds the threshold. 

%

We note that our approach for switching is different from that of  \cite{wang2017optimal}, which makes the switch for a specific context-action pair if the corresponding importance weight is too large. The proposed KL divergence based approach has an averaging effect on the individual weights since it is a weighted sum of the importance weights on a logarithmic scale and the thresholding is context-specific, rather than context-action pair specific. So, the chosen model is either the DM model or the DR model for the specific context and {\it all} possible actions. From a practical standpoint, the thresholding for the KL divergence based approach can be performed on a more compact grid than an importance weight based switching  since the KL divergence uses the importance weights on a  logarithmic scale, enabling easier optimization. We elaborate on the qualitative differences between the two techniques using a toy example in Section \ref{sec:comparison}.

\subsection{Optimizing the Threshold Parameter}
One important part of implementing the DR-IC estimator is to find the optimal threshold parameter.  We follow an  approach similar to the one  used in \cite{wang2017optimal} to select a $\tau$ that minimizes the MSE of the resulting estimator.

Let $Y_i(\tau)$ denotes the estimated reward given the context $x_i$ for a switching threshold $\tau$. That is,
\begin{align*}
Y_i(\tau)&= (r_i - \widehat{r}_\mathrm{IB}(x_i,a_i))  \frac{\pi (a_i \mid x_i)}{\mu (a_i \mid x_i)}\,\indic(D_\mathrm{KL}(x_i) <\tau) \\
&+  \sum_{a \in \mathcal{A}}  \widehat{r}_\mathrm{IB}(x_i, a)\pi( a \mid x_i)\,\indic(D_\mathrm{KL}(x_i)
< \tau) \\&+ \sum_{a \in \mathcal{A}} \widehat{r}_\mathrm{IB}(x_i, a)\pi( a \mid x_i)\,\indic(D_\mathrm{KL}(x_i)
\geq \tau).
\end{align*}
Since $V_\mathrm{DR\mbox{-}IC}(\tau)=\sum_{i=1}^n Y_i(\tau)/n = \overline{Y}(\tau)$, we have the estimate of the variance, $\widehat{\mathrm{Var}}(V_\mathrm{DR\mbox{-}IC}(\tau))$ as 
\small
\begin{align*}
\widehat{\mathrm{Var}}(V_\mathrm{DR\mbox{-}IC}(\tau))=\mathrm{Var} \left( \overline{Y}(\tau)\right)= \frac{1}{n^2}\sum_{i=1}^n {(Y_i(\tau) - \overline{Y}(\tau))}^2.
\end{align*}
\normalsize
Next, we utilize the fact that $\widehat{V}^{\pi}_\mathrm{IPS}$ as defined in \eqref{eq:IPS-estimator} is an unbiased estimator of $V^{\pi}$ to obtain an estimate of squared bias as a function of the threshold $\tau$, given by  $\widetilde{\mathrm{Bias}}^2(\tau)=(\widehat{V}^{\pi}_\mathrm{DR\mbox{-}IC}(\tau)- \widehat{V}^{\pi}_\mathrm{IPS})^2$. To see this, note that 
\begin{align*}
{\mathrm{Bias}}^2 &= (\mathbb{E}[\widehat{V}^{\pi}_\mathrm{DR\mbox{-}IC}(\tau)] - V^{\pi})^2 \\
&=(\mathbb{E}[\widehat{V}^{\pi}_\mathrm{DR\mbox{-}IC}(\tau)] - \mathbb{E}[\widehat{V}^{\pi}_\mathrm{IPS}] + \mathbb{E}[\widehat{V}^{\pi}_\mathrm{IPS}] - V^{\pi})^2\\
&= (\mathbb{E}[\widehat{V}^{\pi}_\mathrm{DR\mbox{-}IC}(\tau)- \mathbb{E}[\widehat{V}^{\pi}_\mathrm{IPS}] )^2 + (\mathbb{E}[\widehat{V}^{\pi}_\mathrm{IPS}]  - V^{\pi})^2 \\
&+ 2 (\mathbb{E}[\widehat{V}^{\pi}_\mathrm{DR\mbox{-}IC}(\tau)]- \mathbb{E}[\widehat{V}^{\pi}_\mathrm{IPS}] ) (\mathbb{E}[\widehat{V}^{\pi}_\mathrm{IPS}]  - V^{\pi})\\
& \stackrel{(a)}{\approx} (\mathbb{E}[\widehat{V}^{\pi}_\mathrm{DR\mbox{-}IC}(\tau)]- \mathbb{E}[\widehat{V}^{\pi}_\mathrm{IPS}] )^2  \\
& \stackrel{(b)}{\approx}  {\left(\frac{1}{n} \sum_{i=1}^n Y_i(\tau) - \frac{1}{n} \sum_{i=1}^n  \frac{\pi(a_i \mid x_i)}{\mu(a_i \mid x_i)}r_i(x_i, a_i)\right)}^2\\
&= (\widehat{V}^{\pi}_\mathrm{DR\mbox{-}IC}(\tau)- \widehat{V}^{\pi}_\mathrm{IPS})^2 = \widetilde{\mathrm{Bias}}^2(\tau).
\end{align*}
Here, to get  $(a)$,  we have used the fact that $(\mathbb{E}[\widehat{V}^{\pi}_\mathrm{IPS}]  - V^{\pi})$ goes to zero as we get more samples, since $\widehat{V}^{\pi}_\mathrm{IPS}$ is an unbiased estimator of $V^{\pi}$. To get $(b)$, we replace the expectation by sample average.

However, due to possibly large importance weights, $\widehat{V}^{\pi}_\mathrm{IPS}$ can be a poor estimate of the true expected reward $V^{\pi}$. We overcome this issue as follows. Let ${R}_{\max} = \max_{x, a} r(x,a)$.  We then propose the following  upperbound of the bias, $\mathrm{BiasUB}(\tau)$:
\begin{align*}
\mathrm{BiasUB}(\tau)&=\frac{1}{n}\sum_{i=1}^n \sum_{a \in \mathcal{A}} R_{\max} \pi (a \mid x_i)\, \indic(D_\mathrm{KL}(x_i) \geq \tau).
\end{align*}
Finally, we will use the following as  our estimate for the squared bias:
\begin{align*}
\widehat{\mathrm{Bias}}^2(\tau)={\min}\left(\widetilde{\mathrm{Bias}}^2(\tau), \mathrm{BiasUB}^2(\tau)\right).
\end{align*}
With these estimates, we optimize the threshold paramater  $\tau$ as
\begin{align}
\label{eq:tau-hat}
\widehat{\tau} = \underset{\tau}{\argmin} \left [ \widehat{\mathrm{Bias}}^2(\tau) + \widehat{\mathrm{Var}}(\tau) \right].
\end{align}

\section{Analysis}
\label{sec:analysis}

\subsection{Main Results}
In this section, we investigate the theoretical properties of the information borrowing estimator $\widehat{r}_{\mathrm{IB}}(z)$ for the reward function and {show that it is asymptotically unbiased. More precisely, in Theorem \ref{thm:bias}, we show that   the bias of the modified  DM estimator $\widehat{V}^{\pi}_{\mathrm{DM\mbox{-}IB}}$  as defined in \eqref{eq:V-DM-IB}  goes to zero when $n$ increases to infinity. In addition to this bias analysis, we  also provide a bound on the MSE of the DR-IC estimator $\widehat{V}^{\pi}_{\mathrm{DR\mbox{-}IC}}(\tau)$ in Theorem \ref{thm:mse}}.

In this section, we assume that action $a$ is continuous   for the purpose of deriving general results. We also make the following assumptions about the problem. 
\begin{assumption} 
\label{assump:reward}
(i). The reward function $r(x,a)$ is continuous in $x$ and $a$. \\ 
(ii) The importance weight $w(x,a)={\pi(a\mid x)}/{\mu(a\mid x)}$ is continuous in $x$ and $a$. Also, there exists a positive constant $w_{\max}$ such that $ w(x,a) \in (0, w_{\max}]$ for all $(x, a)$. \\
(iii) The context  distribution $\nu_{C}$ and the reward distribution $\mu_{R}(\cdot | x, a)$ are continuous. 
\end{assumption}
We make the following assumption about the kernel $K(\cdot)$ we are using in \eqref{covf}. 
\begin{assumption} 
\label{assump:kernel}
    Assume the kernel function $K(\cdot)$ is isotropic and satisfies $\int K(u)du=1$, $\int uK(u) du = 0$, $\int u^2K(u) du <\infty$.
\end{assumption}
We also make the following assumption about the least squares regression based estimate used in \eqref{eq:ls-estimate} and \eqref{eq:r-IB-1}. 
\begin{assumption} \label{assump:lse}
Let $\widehat{\theta}_{\mathrm{ls}}=(\widehat{\theta}_x^\top, \widehat{\theta}_a)^\top = (Z^\top Z)^{-1}Z^\top\mathbf{r}$. $\|\widehat{\theta}_x\|_2$ and $\|\widehat{\theta}_a\|_2$ are bounded from above by some positive constant.
\end{assumption} 
The following assumption is common among the problem of regression estimation, see \cite{nadaraya1964estimating} for more details.
\begin{assumption} \label{assump:sigmar}
$(i)$ The value of $\Sigma(z, \mathbf{r})\bm{1}_n$ is $n(1+o(1))$ for every $z=(x^\top,a)^\top$, 
$(ii)$ $\Sigma_\mathbf{r}$ is a diagonal matrix and its diagonal entries are positive constants.
\end{assumption}

We now present our  theorem about the bias of $\widehat{V}^{\pi}_{\mathrm{DM\mbox{-}IB}}$.
\begin{theorem}[Asymptotic unbiasedness of $\widehat{V}^{\pi}_{\mathrm{DM\mbox{-}IB}}$]
\label{thm:bias}
Let Assumptions \ref{assump:reward}--\ref{assump:sigmar}  hold. 
Assume 
the bandwidth $h_n\rightarrow 0$ as $n\rightarrow\infty$. Then, the bias of the modified direct method estimator $\widehat{V}^{\pi}_{\mathrm{DM\mbox{-}IB}}$ as defined in \eqref{eq:V-DM-IB} converges to zero in probability as $n$ goes to infinity.  
\end{theorem}

The DM-IB estimator inherits its asymptotically unbiased property from $\widehat{r}_\mathrm{IB}(z)$. In the following, we provide a brief intuition on the proof technique. The bias of $\widehat{r}_\mathrm{IB}(z)$ can be written as three terms  involving convolutions with the non-stationary kernel (\ref{covf}). As the bandwidth $h_n$ converges to zero with the sample size, the non-stationary kernel becomes degenerate at each sample in the derived data set $\widetilde{\mathcal{D}}$ such that the convolutions converge to the same sample in $\widetilde{\mathcal{D}}$. Hence the convolution approximates the target and results in the elimination of the bias caused by least squares estimate with the information borrowing term in (\ref{eq:r-IB-1}).

\begin{remark}
Since the underlying reward function is non-linear, it is well-known that the bias of its least squares estimate  $z^\top\widehat{\theta}_\mathrm{ls}$ is lower bounded by a constant. Hence, the bias of the traditional direct method estimator $\widehat{V}_\mathrm{DM}$ is lower bounded by a constant as well.
\end{remark}

We now give an upper bound on the MSE for our DR-IC estimator given in \eqref{eq:v-dr-ic}.
\begin{theorem}[Bound on the MSE of $\widehat{V}^{\pi}_{\mathrm{DR\mbox{-}IC}}$]
\label{thm:mse} 
Let Assumption \ref{assump:reward}.(ii) holds. Let $\epsilon_{\mathrm{IB}}(x,a)= \widehat{r}_{\mathrm{IB}}(x,a)-\mathbb{E}[r| x, a]$ and assume that $ \widehat{r}_{\mathrm{IB}}(x,a) \in [0, R_{\max}]$. Then, the MSE of $\widehat{V}^{\pi}_{\mathrm{DR\mbox{-}IC}}$ as given \eqref{eq:v-dr-ic} with the threshold parameter $\tau$ is at most  
\small
\begin{align*}
    & \frac{2}{n} \mathbb{E}_{\mu}\left[\mathbb{E}\big[(r-\widehat{r}_{\mathrm{IB}}(x,a))^2\big| x ,a\big]+w^2_{\max}\epsilon^2_{\mathrm{IB}}(x,a)\right] \\&+ \frac{2}{n}|\mathcal{A}|\mathrm{E}_\pi[R^2_{\max}] + \mathbb{E}_{\pi}\Big[\epsilon_{\mathrm{IB}}(x,a)\indic\{D_\mathrm{KL}(x)<\tau\} \Big]^2,
\end{align*}
\normalsize
\end{theorem}

{Theorem \ref{thm:mse} shows that the upper bound of the MSE of our DR-IC estimator is at most $O(1/n)$. A similar result of the switch-DR estimator can be found in \cite{wang2017optimal} where the upper bound is $O(1/n)$ as well.} So, in terms of the MSE, our proposed DR-IC estimator is at least as good as the switching method proposed in \cite{wang2017optimal}. At the same time, our approach is better in terms of the bias, as shown in Theorem \ref{thm:bias}.

\subsection{Comparing Different Switch Estimators}
\label{sec:comparison}

The idea of combining two different estimators in off-policy evaluation has been explored in \cite{wang2017optimal}, where the authors proposed a switch estimator (switch-DR) which can switch between DM and DR (or IPS) depending on the magnitude of importance weights. The DR-IC estimator introduced in this paper is  different from the switch-DR estimator, though it is also designed to switch between DM and DR. The switching criterion of the DR-IC estimator is the KL divergence between the logging policy and the target policy, given any context. This measures the overall closeness between the two policies, while the switch-DR estimator uses the importance weights for each individual samples.

\begin{figure}
  \begin{center}
    \includegraphics[width=0.4\textwidth]{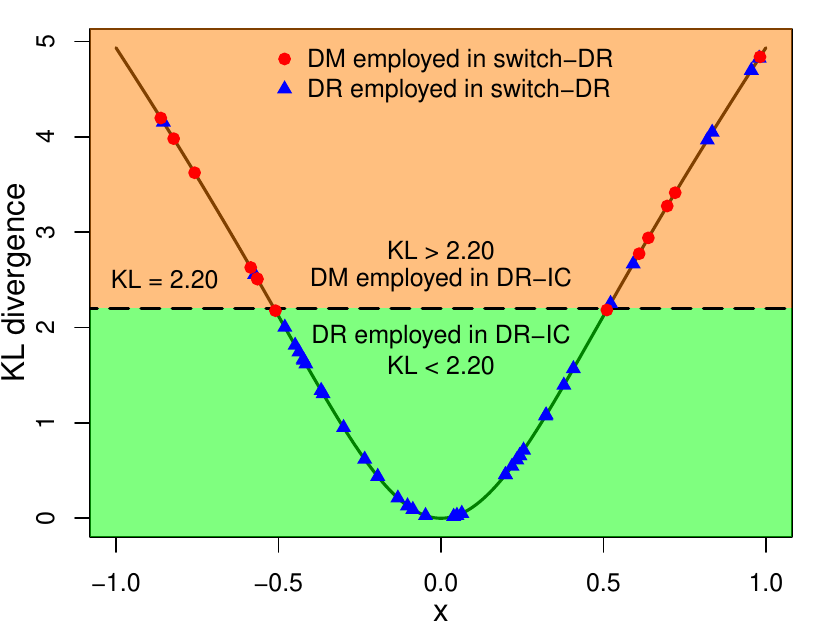}
  \end{center}
  \caption{A toy example for comparing two switching criteria between switch-DR and DR-IC.}
  \label{fig:compare}
\end{figure}

To better demonstrate the differences between these two switching criteria, we consider a toy example designed as follows. Consider a setting with only two actions, denoted as $0, 1$. Assume that the  logging policy has the form $\mu(1\mid x)=e^{5x}/(1+e^{5x})$ and $\mu(0\mid x) = 1- \mu(1\mid x)$. Also, consider the  evaluation policy of the form $\pi(1\mid x)=e^{-5x}/(1+e^{-5x})$ and $\pi(0\mid x) = 1- \pi(1\mid x)$. The contexts are uniformly distributed between $-1$ and $1$. We generate 50 training samples and 50 test samples. The threshold of importance weights for switch-DR is 12.65 and the threshold of KL divergences for DR-IC is 2.20. We plot the KL divergences as a function of contexts in Fig.~\ref{fig:compare} and all 50 samples are marked with red circles or blue triangles based on which method (DM or DR) is employed in the switch-DR estimator. The plot are split into two parts by the threshold of KL divergences in the DR-IC estimator. It is obvious that two switching criteria do not agree with each other always. Some contexts with smaller importance weights where DR is employed in the switch-DR estimator have larger KL divergences, such that DM is employed in the DR-IC estimator.

\section{Experiments}
\label{sec:experiments}

\begin{figure*}[ht]
    \centering
	\includegraphics[width=\linewidth]{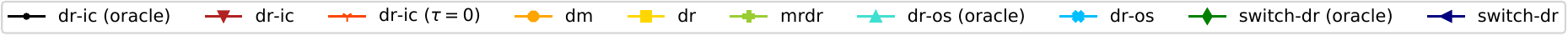}
	\centering
	\begin{minipage}{.24\textwidth}
		\centering
		\includegraphics[width=\linewidth]{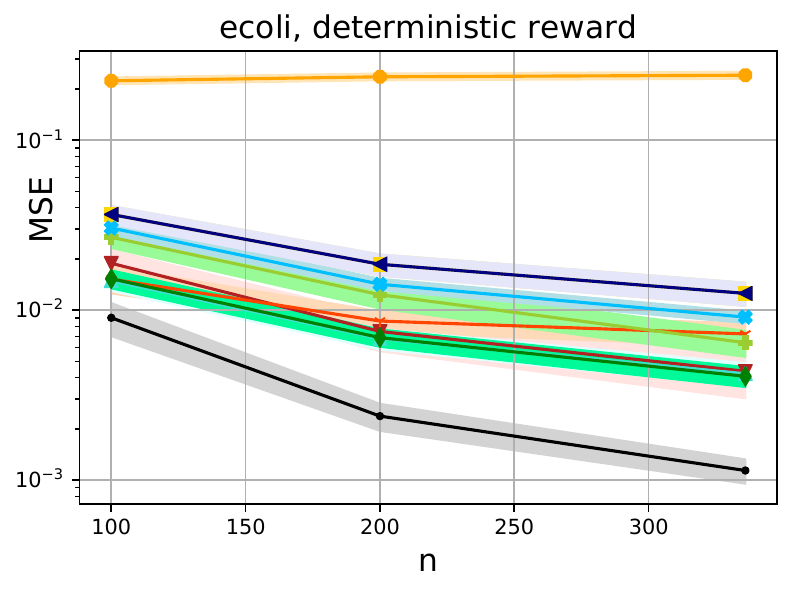}
	\end{minipage}
	\begin{minipage}{.24\textwidth}
		\centering
		\includegraphics[width=\linewidth]{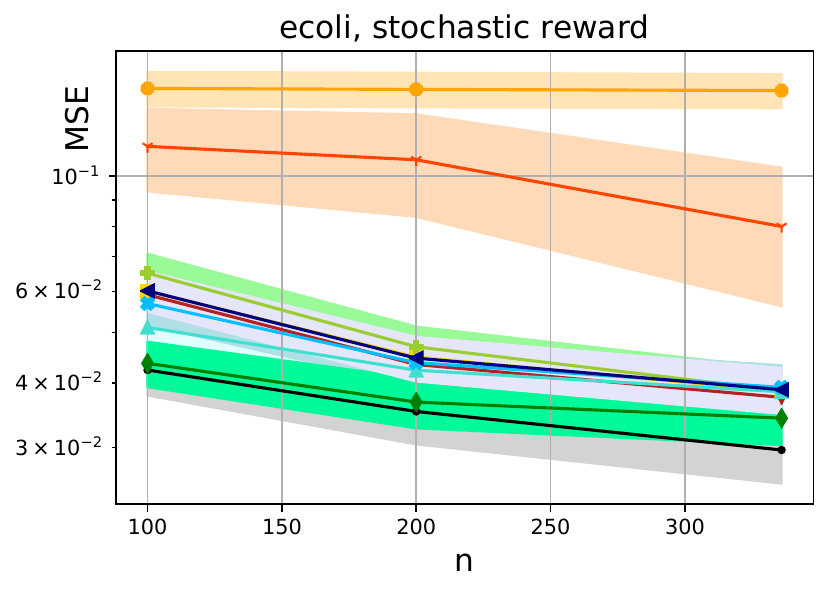}
	\end{minipage}
	\begin{minipage}{.24\textwidth}
		\centering
		\includegraphics[width=\linewidth]{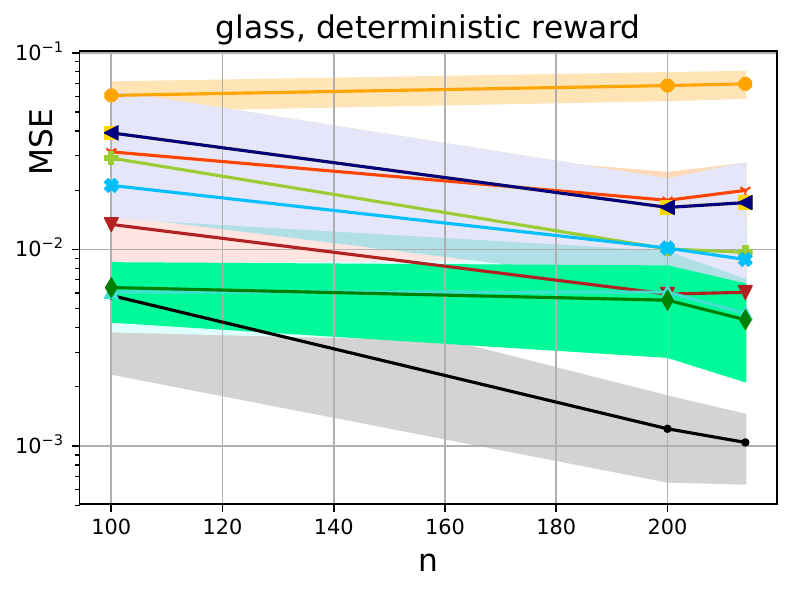}
	\end{minipage}
	\begin{minipage}{.24\textwidth}
		\centering
		\includegraphics[width=\linewidth]{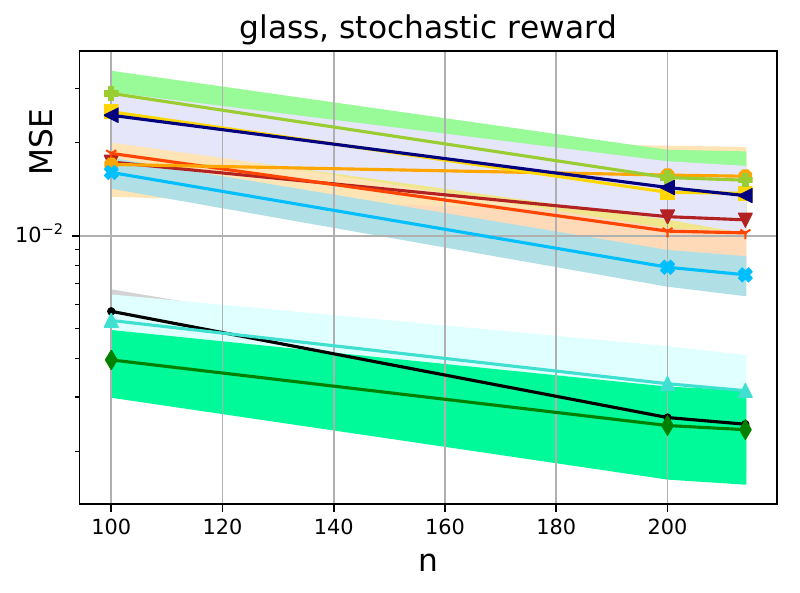}
	\end{minipage}
	\centering
	\begin{minipage}{.24\textwidth}
		\centering
		\includegraphics[width=\linewidth]{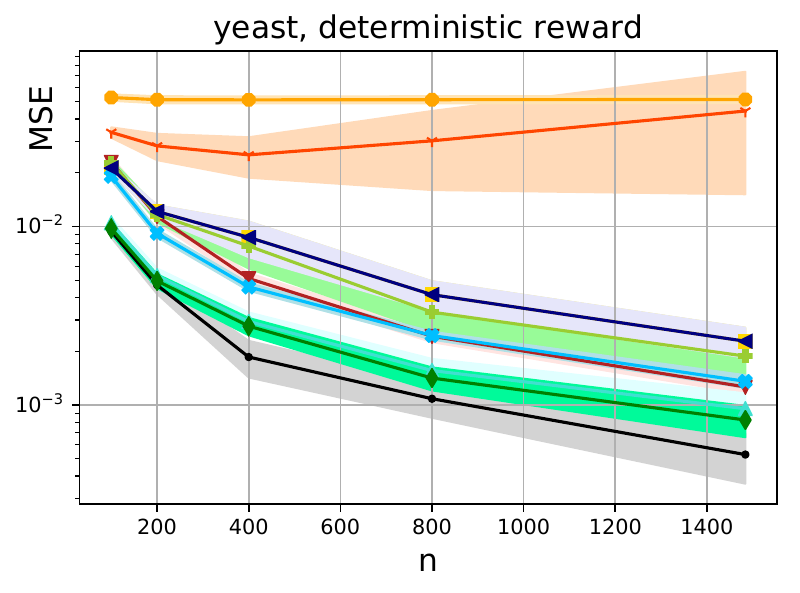}
	\end{minipage}
	\begin{minipage}{.24\textwidth}
		\centering
		\includegraphics[width=\linewidth]{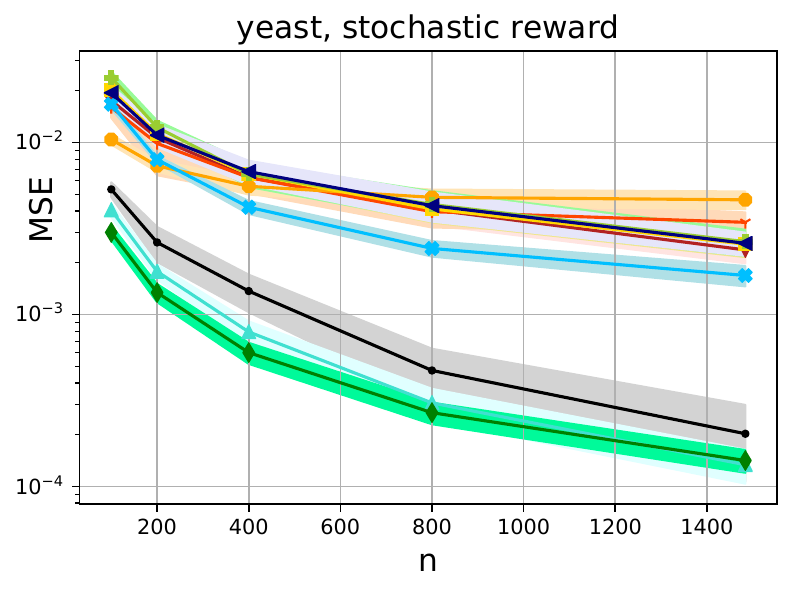}
	\end{minipage}
	\begin{minipage}{.24\textwidth}
		\centering
		\includegraphics[width=\linewidth]{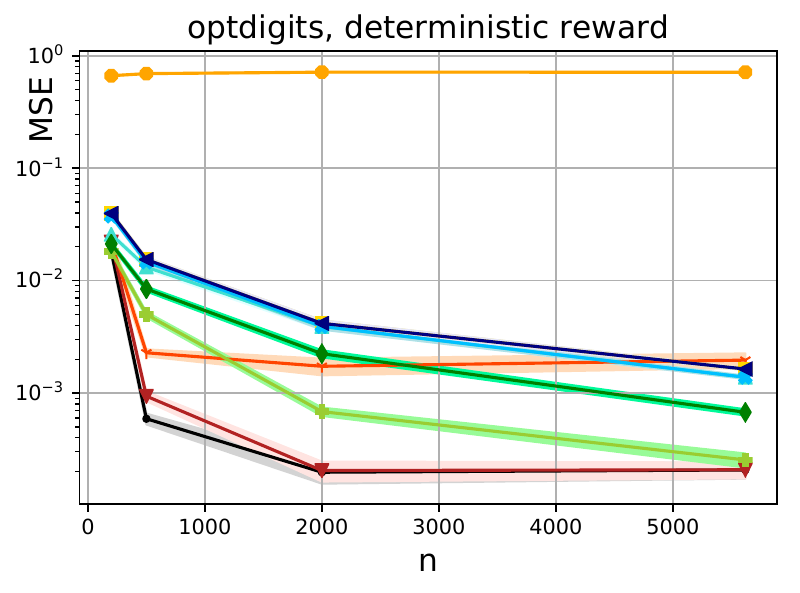}
	\end{minipage}
	\begin{minipage}{.24\textwidth}
		\centering
		\includegraphics[width=\linewidth]{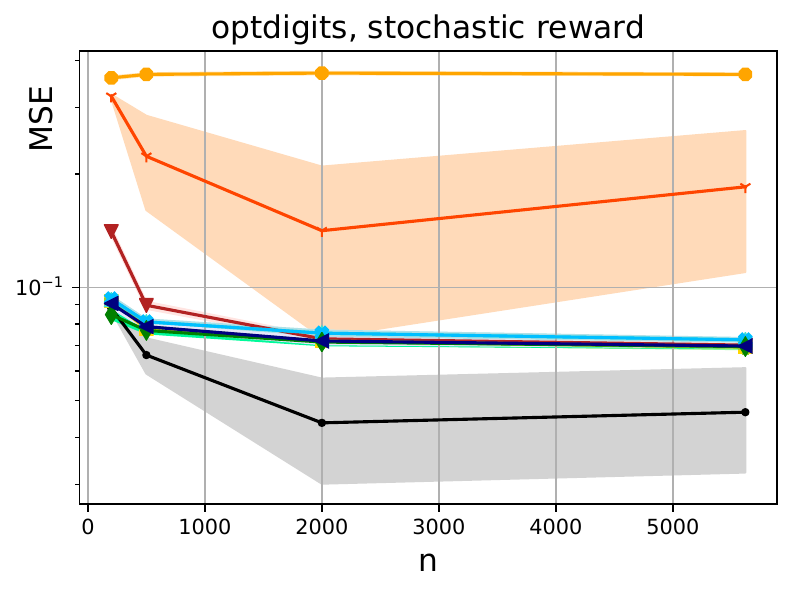}
	\end{minipage}
	\captionof{figure}{\textit{Off-policy evaluation simulation results.} MSE versus sample sizes $n$ for two different reward distributions are showcased for \textit{ecoli, glass, yeast, and optdigits} UCI datasets.}
	\label{figcomp}
\end{figure*}

In this section, we outline our experimental results built on the Open Bandit Pipeline \citep{saito2020open} library.  For the choice of datasets, we follow \citep{dudik2011doubly, dudik2014doubly, wang2017optimal, farajtabarCG18, su2020doubly} to transform the UC Irvine (UCI) multi-class classification datasets \citep{dua2019uci} into off-policy evaluation datasets. The notion of classes in a multi-class dataset becomes the action space in a contextual bandit problem, thus the predicted label becomes the action taken. The classification error becomes the observed loss $l$. Note that this is equivalent to observing a reward $r$, under the transformation $r=1-l$. Specifically, we consider the indicator loss. That is,  for a multi-class example $(x,a^*)$, whenever the rewards are deterministic, then taking an action $a$ yields the reward $r = \indic\{a = a^*\}$, 
or whenever the rewards are stochastic, then taking an action $a$ yields the reward $r = \indic\{a = a^*\}$ with probability $0.7$ and $r = 1-\indic\{a = a^*\}$ otherwise.

We split every dataset into two parts - a training set (70\%) and a test set (30\%). We generate a context-dependent logging policy based on the training set, and use this logging policy to generate the actions. Towards this end, we generate actions uniformly at random, and train a logistic regression model on this training set with randomized actions. We use this model on the test set to get the predicted action distribution that is used as a logging policy. This logging policy $\mu$ then is used to generate $n$ sized bandit data by sampling a context $x$ from the entire dataset, sampling an action $a \sim \mu(\cdot \mid x)$, and then observing a deterministic or stochastic reward $r$. The value of $n$ varies across different datasets.

We also use this training set to generate an evaluation policy by repeating the procedure of getting a logging policy, with the only difference is that the logistic regression model is trained with the true actions present in the training set. We use this evaluation policy $\pi$ to predict the actions in the testing set, and calculate the expected reward, which forms the ground truth value for the given UCI dataset.

We measure the performance of various estimators with clipped mean squared error (MSE), i.e. $\mathbb{E}[(\widehat{V}^{\pi} - V^{\pi})^2 \wedge 1]$ where $\widehat{V}^{\pi}$ is the value of the estimator and $V^{\pi}$ is the ground truth. We use $300-500$ replicates of bandit data generation to estimate the MSE and repeat this process for $10-40$ different seeds to generate the bands (which is $\pm0.5$ the standard deviation) around the estimated MSE.

We compare the MSE of the following estimators. $1.$ DR-IC: our proposed estimator given in \eqref{eq:v-dr-ic} with the tuning parameters optimized  according to \eqref{eq:tau-hat}, $2.$ DR-IC (oracle): our proposed estimator with the tuning parameter optimized using the actual MSE with respect to the ground truth,   $3.$ DR-IC($\tau=0$): this is our proposed  $\widehat{V}^{\pi}_{\mathrm{DM\mbox{-}IB}}$ as given in \eqref{eq:V-DM-IB},  $4.$ DM: as given \eqref{eq:DM-estimator} using traditional ridge regression reward estimate \eqref{eq:ls-estimate},  $5.$ DR: standard DR estimator according to \eqref{eq:DR-Estimator} using the reward estimate \eqref{eq:ls-estimate}, $6.$ MR-DR: More robust Doubly robust estimator based on \cite{farajtabarCG18}, $7.$ DR-OS (oracle): the oracle tuned version of the shrinkage based estimator proposed in \cite{su2020doubly} , $8.$ DR-OS: optimized version of the shrinkage estimator proposed in \cite{su2020doubly}, $9.$ switch-DR (oracle): the oracle version of the switch estimator proposed in \cite{wang2017optimal}, and $10.$ switch-DR : the switch estimator proposed in \cite{wang2017optimal} with tuning parameters optimized using an estimated MSE.
In Appendix D, we include a separate simulation study comparing DR-IC($\tau=0$) with a standard Nadaraya Watson estimator, to highlight the benefits of the specific kernel introduced in our model, in effectively borrowing information. 

In Fig~\ref{figcomp}, we  show several attractive properties of the proposed method with respect to the state of the art algorithms. First, in seven out of the eight cases considered, the borrowing of information displays a significant practical improvement as shown in the performance of DR-IC($\tau=0$) (DM-IB) over the standard DM across all sample sizes. In fact, in five out of the eight cases, DR-IC ($\tau=0$) either outperforms or is as good as a traditional DR approach across the different sample sizes. 

Second, the proposed context-based switching algorithm improves upon the performance of DR-IC($\tau=0$) and clearly dominates the state-of-the-art methods in seven out of the eight cases considered. The oracle version of the proposed  method DR-IC (oracle) significantly outperforms the oracle versions of other methods, while the tuned version DR-IC is either as good as or better than the tuned versions of the competitors. This also highlights the advantages of the KL based switching approach over the importance weight based switching approach of \cite{wang2017optimal}. In five out of the eight cases, DR-IC shows significant improvement over switch-DR, and is as good as switch-DR in the remaining three cases. 

Third, the  results indicate that for the deterministic reward model, the DR-IC and DR-IC(oracle) are at least as good as, or significantly better than all their counterparts. For the stochastic reward model, the variability increases for all estimators. 
Nonetheless, DR-IC is still atleast as good as, or better than all the other estimators  in three out of four cases.



\textbf{Additional experiments:} We note that we have included more experiment results and details of the implementation  in the appendix. In Appendix D, we have provided a comparison of the proposed approach with a kernel based non-parametric reward estimator, and illustrate the advantages of the proposed formulation. We have also included the code in this \href{https://github.com/kishanpb/OffPolicyEvaluation_InformationBorrowing_ContextSwitching}{github page}.

\section{Conclusions}
\label{sec:conclusions}
In this work,  we have introduced the idea of information borrowing which utilizes the similarity structure among context-action pairs and the importance weights associated with the logging and evaluation policies to significantly improve the practical properties of a traditional direct method for OPE. Under assumptions discussed in Section \ref{sec:analysis}, the proposed reward estimator is asymptoticallly unbiased, which is usually not the case for traditional DM reward models. We further introduce a KL divergence  based switching algorithm that employs the strengths of information borrowing,  resulting in a context-based switch estimator that dominates the state of the art algorithms in most cases. These advantages are relevant not just for the fundamental problem of off-policy evaluation as considered in this article, but for more general problems in off-policy optimization and reinforcement learning as well. The notion of effective information borrowing becomes particularly crucial while assimilating information from multiple context distributions  or logging policies with varying similarities to the evaluation policy. These problems indicate exciting possibilities for future research using the idea of effective policy dependent information borrowing introduced in this paper. Another exciting direction of research would be to identify better optimization schemes for the tuning parameter $\tau$ and get the tuned estimate closer to the oracle version.


\bibliography{OffPolicy-References}

\onecolumn
\appendix
\allowdisplaybreaks

\section{Proof of Theorem \ref{thm:bias}}
For simplicity, we prove Theorem \ref{thm:bias} when $x$ and $a$ are scalars and we assume $\nu_C(x)=U(0,1)$ and $\mu(a\mid x) = U(0,1)$, where $U(0,1)$ is the uniform distribution between 0 and 1 with density function $\mathbf{1}_{(0,1)}(\cdot)$. In addition, we assume there is no intercept term in the lest square estimate  $\widehat{\theta}_\mathrm{ls}$. For more general scenarios, some modifications of the assumptions may be necessary. The kernel function is
\begin{align*}
    [\Sigma(\widetilde{\mathbf{r}}, \mathbf{r})]_{ji} & = \frac{1}{h_n\sqrt{\widetilde{w}_jw_i}}K\Big(\frac{\widetilde{x}_j-x_i}{h_n \sqrt{\widetilde{w}_j w_i}}\Big)\frac{1}{h_n\sqrt{\widetilde{w}_jw_i}}K\Big(\frac{\widetilde{a}_j-a_i}{h_n \sqrt{\widetilde{w}_j w_i}}\Big).
\end{align*}

\textbf{Main idea of the proof:} We first show that the information borrowing estimator $\widehat{r}_{\mathrm{IB}}(z)$ evaluated at any point $z=(x,a)^\top$ is asymptotically unbiased. This essentially leads to the asymptotic unbiasedness of the direct method estimator with $\widehat{r}_{\mathrm{IB}}(z)$, $\widehat{V}_\mathrm{DM-IB}$. The proof follows three main steps: 

\textbf{(i).} We decompose $\widehat{r}_{\mathrm{IB}}(z)$ into two parts: a weighted sum of the least squared estimate $\widehat{\theta}_\mathrm{ls}$, i.e., the first two terms in (\ref{eq:rhatIB}), and a weighted sum of $\mathbf{r}$, i.e., the third term in (\ref{eq:rhatIB}). Hence, its expectation with respect to $\nu_R(r|x,a)$ follows the same construction where only $\widehat{\theta}_\mathrm{ls}$ and $\mathbf{r}$ are replaced by their expectations, see (\ref{eq:exprIB}). 

\textbf{(ii).} We show $A_n\rightarrow x, B_n\rightarrow a$ and $C_n\rightarrow r(x,a)$ for any $(x,a)$. They can be seen as three convolutions with the non-stationary kernel (\ref{covf}).
Due to the asymptotic degeneracy of the non-stationary kernel when the bandwidth $h_n$ goes to zero with the sample size, all the three convolutions converge to its corresponding degenerate point $x$, $a$, $r(x,a)$, respectively.

\textbf{(iii).} As  $\widehat{\theta}_\mathrm{ls}$ is norm-bounded, the first two terms in (\ref{eq:exprIB}) converge to zero and the last term converges to the reward function at $r(x,a)$ at $(x,a)$. These observations lead to $\mathbb{E}_{r\mid x,a}[\widehat{r}_{\mathrm{IB}}(z)\mid \bm{x},\bm{a}]\stackrel{p}{\rightarrow} r(x,a)$. and subsequently the asymptotic unbiasedness of $\widehat{V}_\mathrm{DM-IB}$.

The information borrowing estimator $\widehat{r}_{\mathrm{IB}}(z)$ for the rewards function $r(x,a)$ can be decomposed into three terms defined as follows.

\small
\begin{align}
    \widehat{r}_{\mathrm{IB}}(z) & = z^\top\widehat{\theta}_{\mathrm{ls}} + \Sigma({z,\mathbf{r}})\Sigma^{-1}_\mathbf{r}[\mathrm{diag}\{\Sigma(z,\bm{r})\Sigma^{-1}_\mathbf{r}\bm{1}_n\} ]^{-1}(\mathbf{r} - Z\widehat{\theta}_{\mathrm{ls}}) \nonumber\\
    & = \bigg(z - \frac{1}{n(1+o(1))}\Sigma(z,\mathbf{r})Z\bigg ) \widehat{\theta}_{\mathrm{ls}} + \frac{1}{n(1+o(1))}\Sigma(z,\mathbf{r})\,\mathbf{r},  \qquad \text{ by Assumption \ref{assump:sigmar}}  \nonumber\\
    & = \bigg(x-\frac{1}{n(1+o(1))}\sum_{i=1}^n\frac{1}{h_n\sqrt{w_zw_i}}K\Big(\frac{x-x_i}{h_n \sqrt{w_z w_i}}\Big)\frac{1}{h_n\sqrt{w_zw_i}}K\Big(\frac{a-a_i}{h_n \sqrt{w_z w_i}}\Big)\,x_i\bigg)\widehat{\theta}_x \nonumber\\
    & \phantom{1111111111}+ \bigg(a-\frac{1}{n(1+o(1))}\sum_{i=1}^n\frac{1}{h_n\sqrt{w_zw_i}}K\Big(\frac{x-x_i}{h_n \sqrt{w_z w_i}}\Big)\frac{1}{h_n\sqrt{w_zw_i}}K\Big(\frac{a-a_i}{h_n \sqrt{w_z w_i}}\Big)\,a_i\bigg)\widehat{\theta}_a \nonumber\\ 
    & \phantom{11111111111111111111}+ \frac{1}{n(1+o(1))} \sum_{i=1}^n \frac{1}{h_n\sqrt{w_zw_i}}K\Big(\frac{x-x_i}{h_n \sqrt{w_z w_i}}\Big)\frac{1}{h_n\sqrt{w_zw_i}}K\Big(\frac{a-a_i}{h_n \sqrt{w_z w_i}}\Big)\,r_i, \label{eq:rhatIB}
\end{align}
\normalsize
where $\bm{x}=(x_1,\ldots,x_n)^\top, \bm{a}=(a_1,\ldots,a_n)^\top$ and $
    w_z = w(x,a) = \frac{\pi(a\mid x)}{\mu(a\mid x)},  w_i = w(x_i,a_i) = \frac{\pi(a_i\mid x_i)}{\mu(a_i\mid x_i)}.$ Now, 
\small
\begin{align}
    \mathbb{E}_{r\mid x,a}[\widehat{r}_{\mathrm{IB}}(z)\mid \bm{x},\bm{a}] 
    & = \bigg\{x-\frac{1}{n(1+o(1))}\sum_{i=1}^n\frac{1}{h_n\sqrt{w_zw_i}}K\Big(\frac{x-x_i}{h_n \sqrt{w_z w_i}}\Big)\frac{1}{h_n\sqrt{w_zw_i}}K\Big(\frac{a-a_i}{h_n \sqrt{w_z w_i}}\Big)\,x_i\bigg\}\cdot\mathbb{E}_r[\widehat{\theta}_x\mid \bm{x},\bm{a}] \nonumber\\
    & \phantom{11111}+ \bigg\{a-\frac{1}{n(1+o(1))}\sum_{i=1}^n\frac{1}{h_n\sqrt{w_zw_i}}K\Big(\frac{x-x_i}{h_n \sqrt{w_z w_i}}\Big)\frac{1}{h_n\sqrt{w_zw_i}}K\Big(\frac{a-a_i}{h_n \sqrt{w_z w_i}}\Big)\,a_i\bigg\}\cdot\mathbb{E}_r[\widehat{\theta}_a\mid \bm{x},\bm{a}] \nonumber\\
    & \phantom{1111111111} + \frac{1}{n(1+o(1))} \sum_{i=1}^n \frac{1}{h_n\sqrt{w_zw_i}}K\Big(\frac{x-x_i}{h_n \sqrt{w_z w_i}}\Big)\frac{1}{h_n\sqrt{w_zw_i}}K\Big(\frac{a-a_i}{h_n \sqrt{w_z w_i}}\Big)\cdot r(x_i,a_i) \nonumber\\
    &:=(x-A_n)\cdot\mathbb{E}_r[\widehat{\theta}_x\mid \bm{x},\bm{a}] + (a-B_n)\cdot\mathbb{E}_r[\widehat{\theta}_a\mid \bm{x},\bm{a}] + C_n. \label{eq:exprIB}
\end{align}
\normalsize
Next we show $A_n \stackrel{p}{\rightarrow} x$ and $B_n \stackrel{p}{\rightarrow} a$, where $\stackrel{p}{\rightarrow}$ represents convergence in probability.
\small
\begin{align*}
    \mathbb{E}_{x,a}[A_n]
    & = \frac{1}{n(1+o(1))}\sum_{i=1}^n\int_{a_i}\int_{x_i}\frac{1}{h_n\sqrt{w_zw_i}}K\Big(\frac{x-x_i}{h_n \sqrt{w_z w_i}}\Big)\frac{1}{h_n\sqrt{w_zw_i}}K\Big(\frac{a-a_i}{h_n \sqrt{w_z w_i}}\Big)\,x_i\bm{1}_{(0,1)}(x_i)\bm{1}_{(0,1)}(a_i) \,dx_i\,da_i \\
    & = \int_{-\infty}^{+\infty}\Bigg\{\int_{-\infty}^{+\infty}\frac{1}{1+o(1)}\frac{1}{h_n\sqrt{w_zw_{z'}}}K\Big(\frac{x-x'}{h_n \sqrt{w_z w_{z'} }}\Big)\frac{1}{h_n\sqrt{w_zw_{z'}}}K\Big(\frac{a-a'}{h_n \sqrt{w_z w_{z'}}}\Big)\,x'\bm{1}_{(0,1)}(x')\,dx'\Bigg\}\bm{1}_{(0,1)}(a')\,da'\\
    & = \Bigg(\int_{a-c_nh_n\sqrt{w_z}}^{a+c_nh_n\sqrt{w_z}} + \int_{-\infty}^{a-c_nh_n\sqrt{w_z}} + \int_{a+c_nh_n\sqrt{w_z}}^{+\infty}\Bigg)\Bigg(\int_{x-c_nh_n\sqrt{w_z}}^{x+c_nh_n\sqrt{w_z}} + \int_{-\infty}^{x-c_nh_n\sqrt{w_z}} + \int_{x+c_nh_n\sqrt{w_z}}^{+\infty}\Bigg) \\
    & \phantom{1111111111} \frac{1}{1+o(1)}\frac{1}{h_n\sqrt{w_zw_{z'}}}K\Big(\frac{x-x'}{h_n \sqrt{w_z w_{z'} }}\Big)\frac{1}{h_n\sqrt{w_zw_{z'}}}K\Big(\frac{a-a'}{h_n \sqrt{w_z w_{z'}}}\Big)\,x'\bm{1}_{(0,1)}(x')\bm{1}_{(0,1)}(a')\,dx'\,da' \\
    & = \Bigg(\int_{a-c_nh_n\sqrt{w_z}}^{a+c_nh_n\sqrt{w_z}}\int_{x-c_nh_n\sqrt{w_z}}^{x+c_nh_n\sqrt{w_z}} 
    + \int_{a-c_nh_n\sqrt{w_z}}^{a+c_nh_n\sqrt{w_z}}\int_{-\infty}^{x-c_nh_n\sqrt{w_z}}
    + \int_{a-c_nh_n\sqrt{w_z}}^{a+c_nh_n\sqrt{w_z}}\int_{x+c_nh_n\sqrt{w_z}}^{+\infty} \\
    & + \int_{-\infty}^{a-c_nh_n\sqrt{w_z}} \int_{x-c_nh_n\sqrt{w_z}}^{x+c_nh_n\sqrt{w_z}} 
    + \int_{-\infty}^{a-c_nh_n\sqrt{w_z}} \int_{-\infty}^{x-c_nh_n\sqrt{w_z}}
    + \int_{-\infty}^{a-c_nh_n\sqrt{w_z}} \int_{x+c_nh_n\sqrt{w_z}}^{+\infty} \\
    & + \int_{a+c_nh_n\sqrt{w_z}}^{+\infty} \int_{x-c_nh_n\sqrt{w_z}}^{x+c_nh_n\sqrt{w_z}} 
    + \int_{a+c_nh_n\sqrt{w_z}}^{+\infty} \int_{-\infty}^{x-c_nh_n\sqrt{w_z}}
    + \int_{a+c_nh_n\sqrt{w_z}}^{+\infty} \int_{x+c_nh_n\sqrt{w_z}}^{+\infty} \Bigg) \\
    & \phantom{1111111111} \frac{1}{1+o(1)}\frac{1}{h_n\sqrt{w_zw_{z'}}}K\Big(\frac{x-x'}{h_n \sqrt{w_z w_{z'} }}\Big)\frac{1}{h_n\sqrt{w_zw_{z'}}}K\Big(\frac{a-a'}{h_n \sqrt{w_z w_{z'}}}\Big)\,x'\bm{1}_{(0,1)}(x')\bm{1}_{(0,1)}(a')\,dx'\,da' \\
    & := A_{1n} + A_{2n} + A_{3n} + A_{4n} + A_{5n} + A_{6n} + A_{7n} + A_{8n} + A_{9n},
\end{align*}
\normalsize
where $c_n\rightarrow+\infty$, $h_n\rightarrow 0$, and $c_nh_n\rightarrow 0$ as $n\rightarrow\infty$ and $
    z'=(x',a'),  w_{z'} = w(x',a') = \frac{\pi(a'\mid x')}{\mu(a'\mid x')}.$
\normalsize
\small
\begin{align*}
    A_{1n} & = \int_{a-c_nh_n\sqrt{w_z}}^{a+c_nh_n\sqrt{w_z}}\int_{x-c_nh_n\sqrt{w_z}}^{x+c_nh_n\sqrt{w_z}}\frac{1}{1+o(1)}\frac{1}{h_n\sqrt{w_zw_{z'}}}K\Big(\frac{x-x'}{h_n \sqrt{w_z w_{z'} }}\Big)\frac{1}{h_n\sqrt{w_zw_{z'}}}K\Big(\frac{a-a'}{h_n \sqrt{w_z w_{z'}}}\Big)\, \\
    &\phantom{111111111111111111111111111111111111111111111111111111111111111111111111} \cdot x'\bm{1}_{(0,1)}(x')\bm{1}_{(0,1)}(a')\,dx'\,da' \\
    & \text{let } t = \frac{x'-x}{h_n\sqrt{w_z}} \text{ and } s = \frac{a'-a}{h_n\sqrt{w_z}} \\
    & = \int_{-c_n}^{c_n}\int_{-c_n}^{c_n}\frac{1}{1+o(1)}\frac{1}{\sqrt{w_{z'}}}K\Big(\frac{t}{\sqrt{w_{z'}}}\Big)\frac{1}{\sqrt{w_{z'}}}K\Big(\frac{s}{\sqrt{w_{z'}}}\Big)(x+th_n\sqrt{w_z}) \\
    & \phantom{11111111111111111111111111111111111111111111111111111} \cdot\bm{1}_{( -\frac{x}{h_n\sqrt{w_z}}, \frac{1-x}{h_n\sqrt{w_z}} )}(t)\bm{1}_{( -\frac{x}{h_n\sqrt{w_z}}, \frac{1-x}{h_n\sqrt{w_z}} )}(s)\,dt\,ds \\
    & \text{and } w_{z'} = w(x',a') = w(x+th_n\sqrt{w_z},a+th_n\sqrt{w_z})\rightarrow w(x,a) = w_z, \text{ as } n\rightarrow \infty \\
    & \rightarrow x\int_{-\infty}^{+\infty}\int_{-\infty}^{+\infty}\frac{1}{\sqrt{w_{z}}}K\Big(\frac{t}{\sqrt{w_{z}}}\Big)\frac{1}{\sqrt{w_{z}}}K\Big(\frac{s}{\sqrt{w_{z}}}\Big)\,dt\,ds = x.
\end{align*}
\begin{align*}
    A_{2n} & = \int_{a-c_nh_n\sqrt{w_z}}^{a+c_nh_n\sqrt{w_z}}\int_{-\infty}^{x-c_nh_n\sqrt{w_z}}\frac{1}{1+o(1)}\frac{1}{h_n\sqrt{w_zw_{z'}}}K\Big(\frac{x-x'}{h_n \sqrt{w_z w_{z'} }}\Big)\frac{1}{h_n\sqrt{w_zw_{z'}}}K\Big(\frac{a-a'}{h_n \sqrt{w_z w_{z'}}}\Big)\, \\
    &\phantom{111111111111111111111111111111111111111111111111111111111111111111111111} \cdot x'\bm{1}_{(0,1)}(x')\bm{1}_{(0,1)}(a')\,dx'\,da' \\
    & \text{let } t = \frac{x'-x}{h_n\sqrt{w_z}} \text{ and } s = \frac{a'-a}{h_n\sqrt{w_z}} \\
    & = \int_{-c_n}^{c_n}\int_{-\infty}^{-c_n}\frac{1}{1+o(1)}\frac{1}{\sqrt{w_{z'}}}K\Big(\frac{t}{\sqrt{w_{z'}}}\Big)\frac{1}{\sqrt{w_{z'}}}K\Big(\frac{s}{\sqrt{w_{z'}}}\Big)(x+th_n\sqrt{w_z}) \\
    & \phantom{11111111111111111111111111111111111111111111111111111} \cdot \bm{1}_{( -\frac{x}{h_n\sqrt{w_z}}, \frac{1-x}{h_n\sqrt{w_z}} )}(t)\bm{1}_{( -\frac{x}{h_n\sqrt{w_z}}, \frac{1-x}{h_n\sqrt{w_z}} )}(s)\,dt\,ds \\
    & \text{and } w_{z'} = w(x',a') = w(x+th_n\sqrt{w_z},a+th_n\sqrt{w_z})\rightarrow w(x,a) = w_z, \text{ as } n\rightarrow \infty \\
    & \rightarrow x\int_{-\infty}^{+\infty}\int_{-\infty}^{-\infty}\frac{1}{\sqrt{w_{z}}}K\Big(\frac{t}{\sqrt{w_{z}}}\Big)\frac{1}{\sqrt{w_{z}}}K\Big(\frac{s}{\sqrt{w_{z}}}\Big)\,dt\,ds = 0.    
\end{align*}
\normalsize
Similarly, we can show $A_{jn}\rightarrow 0$ as $n\rightarrow \infty$, $j=3,\ldots,9$. Thus, 
\begin{equation*}
    \mathbb{E}_{x,a}[A_n] \rightarrow x, \text{ as } n\rightarrow\infty.
\end{equation*}
Therefore, $A_n \stackrel{p}{\rightarrow} x$ as $n\rightarrow \infty$. Similarly, we can show $B_n \stackrel{p}{\rightarrow} a$ as $n\rightarrow \infty$. Finally, we show $C_n \stackrel{p}{\rightarrow} r(x,a)$.
\small
\begin{align*}
    \mathbb{E}_{x,a}[C_n]
    & = \frac{1}{n(1+o(1))}\sum_{i=1}^n\int_{a_i}\int_{x_i}\frac{1}{h_n\sqrt{w_zw_i}}K\Big(\frac{x-x_i}{h_n \sqrt{w_z w_i}}\Big)\frac{1}{h_n\sqrt{w_zw_i}}K\Big(\frac{a-a_i}{h_n \sqrt{w_z w_i}}\Big)\,\bm{1}_{(0,1)}(x_i)\bm{1}_{(0,1)}(a_i)\,r(x_i,a_i) \,dx_i\,da_i \\
    & = \int_{-\infty}^{+\infty}  \int_{-\infty}^{+\infty}\frac{1}{1+o(1)}\frac{1}{h_n\sqrt{w_zw_{z'}}}K\Big(\frac{x-x'}{h_n \sqrt{w_z w_{z'} }}\Big)\frac{1}{h_n\sqrt{w_zw_{z'}}}K\Big(\frac{a-a'}{h_n \sqrt{w_z w_{z'}}}\Big)\,r(x',a')\,\bm{1}_{(0,1)}(x')\bm{1}_{(0,1)}(a')\,dx'\,da'\\
    & = \Bigg(\int_{a-c_nh_n\sqrt{w_z}}^{a+c_nh_n\sqrt{w_z}}\int_{x-c_nh_n\sqrt{w_z}}^{x+c_nh_n\sqrt{w_z}} 
    + \int_{a-c_nh_n\sqrt{w_z}}^{a+c_nh_n\sqrt{w_z}}\int_{-\infty}^{x-c_nh_n\sqrt{w_z}}
    + \int_{a-c_nh_n\sqrt{w_z}}^{a+c_nh_n\sqrt{w_z}}\int_{x+c_nh_n\sqrt{w_z}}^{+\infty} \\
    & + \int_{-\infty}^{a-c_nh_n\sqrt{w_z}} \int_{x-c_nh_n\sqrt{w_z}}^{x+c_nh_n\sqrt{w_z}} 
    + \int_{-\infty}^{a-c_nh_n\sqrt{w_z}} \int_{-\infty}^{x-c_nh_n\sqrt{w_z}}
    + \int_{-\infty}^{a-c_nh_n\sqrt{w_z}} \int_{x+c_nh_n\sqrt{w_z}}^{+\infty} \\
    & + \int_{a+c_nh_n\sqrt{w_z}}^{+\infty} \int_{x-c_nh_n\sqrt{w_z}}^{x+c_nh_n\sqrt{w_z}} 
    + \int_{a+c_nh_n\sqrt{w_z}}^{+\infty} \int_{-\infty}^{x-c_nh_n\sqrt{w_z}}
    + \int_{a+c_nh_n\sqrt{w_z}}^{+\infty} \int_{x+c_nh_n\sqrt{w_z}}^{+\infty} \Bigg) \\
    & \phantom{1111111111} \frac{1}{1+o(1)}\frac{1}{h_n\sqrt{w_zw_{z'}}}K\Big(\frac{x-x'}{h_n \sqrt{w_z w_{z'} }}\Big)\frac{1}{h_n\sqrt{w_zw_{z'}}}K\Big(\frac{a-a'}{h_n \sqrt{w_z w_{z'}}}\Big)\,r(x',a')\,\bm{1}_{(0,1)}(x')\bm{1}_{(0,1)}(a')\,dx'\,da' \\
    & := C_{1n} + C_{2n} + C_{3n} + C_{4n} + C_{5n} + C_{6n} + C_{7n} + C_{8n} + C_{9n}.
\end{align*}
\begin{align*}
    C_{1n} & = \int_{a-c_nh_n\sqrt{w_z}}^{a+c_nh_n\sqrt{w_z}}\int_{x-c_nh_n\sqrt{w_z}}^{x+c_nh_n\sqrt{w_z}}\frac{r(x',a')}{1+o(1)}\frac{1}{h_n\sqrt{w_zw_{z'}}}K\Big(\frac{x-x'}{h_n \sqrt{w_z w_{z'} }}\Big)\frac{1}{h_n\sqrt{w_zw_{z'}}}K\Big(\frac{a-a'}{h_n \sqrt{w_z w_{z'}}}\Big)\, \\
    &\phantom{111111111111111111111111111111111111111111111111111111111111111111111111} \cdot \bm{1}_{(0,1)}(x')\bm{1}_{(0,1)}(a')\,dx'\,da' \\
    & \text{let } t = \frac{x'-x}{h_n\sqrt{w_z}} \text{ and } s = \frac{a'-a}{h_n\sqrt{w_z}} \\
    & = \int_{-c_n}^{c_n}\int_{-c_n}^{c_n}\frac{r(x+th_n\sqrt{w_z}, a+th_n\sqrt{w_z})}{1+o(1)}\frac{1}{\sqrt{w_{z'}}}K\Big(\frac{t}{\sqrt{w_{z'}}}\Big)\frac{1}{\sqrt{w_{z'}}}K\Big(\frac{s}{\sqrt{w_{z'}}}\Big) \\ 
    & \phantom{11111111111111111111111111111111111111111111111111111} \cdot \bm{1}_{( -\frac{x}{h_n\sqrt{w_z}}, \frac{1-x}{h_n\sqrt{w_z}} )}(t)\bm{1}_{( -\frac{x}{h_n\sqrt{w_z}}, \frac{1-x}{h_n\sqrt{w_z}} )}(s)\,dt\,ds \\
    & w_{z'} = w(x',a') = w(x+th_n\sqrt{w_z},a+th_n\sqrt{w_z})\rightarrow w(x,a) = w_z, \text{ as } n\rightarrow \infty \\
    & r(x+th_n\sqrt{w_z}, a+th_n\sqrt{w_z}) \rightarrow r(x, a) \text{ as } n\rightarrow \infty \\
    & \rightarrow r(x,a)\int_{-\infty}^{+\infty}\int_{-\infty}^{+\infty}\frac{1}{\sqrt{w_{z}}}K\Big(\frac{t}{\sqrt{w_{z}}}\Big)\frac{1}{\sqrt{w_{z}}}K\Big(\frac{s}{\sqrt{w_{z}}}\Big)\,dt\,ds = r(x,a).
\end{align*}
\normalsize
Similarly, we can show $C_{jn}\rightarrow 0$ as $n\rightarrow \infty$, $j=2,\ldots,9$. Thus, 
\begin{equation*}
    \mathbb{E}_{x,a}[C_n] \rightarrow r(x,a), \text{ as } n\rightarrow\infty.
\end{equation*}
Therefore, $C_n \stackrel{p}{\rightarrow} r(x,a)$ as $n\rightarrow \infty$. By Assumption \ref{assump:lse}, $\mathbb{E}_{r\mid x,a}[\widehat{r}_{\mathrm{IB}}(z)\mid \bm{x},\bm{a}] \stackrel{p}{\rightarrow} r(x,a), \text{ as } n\rightarrow \infty$.
\begin{align*}
    \mathbb{E}[\widehat{V}_\mathrm{DM-IB}] & = \mathbb{E}\Big[\frac{1}{n}\sum_{i=1}^n\sum_{a\in \mathcal{A}}\pi(a\mid x_i)\widehat{r}_\mathrm{IB}(x_i,a) \Big] \\
    & = \mathbb{E} \Big[\sum_{a\in \mathcal{A}}\pi(a\mid x)\widehat{r}_\mathrm{IB}(x,a)\Big] \\
    & = \mathbb{E}_{\pi}[\widehat{r}_\mathrm{IB}(x,a)] \\
    & = \mathbb{E}_x\mathbb{E}_{a\mid x\sim \pi}\mathbb{E}_{r\mid x,a}[\widehat{r}_\mathrm{IB}(x,a)]
    \stackrel{p}{\rightarrow} \mathbb{E}_x\mathbb{E}_{a|x\sim\pi}\mathbb{E}_{r\mid x,a}[r(x,a)] = \mathbb{E}_\pi[r] = V^\pi, \quad \text{as } n\rightarrow \infty.
\end{align*}
We conclude that the bias of the direct method estimator with the information borrowing estimator $\widehat{r}_{\mathrm{IB}}(z)$ goes to zero in probability as $n\rightarrow \infty$.


\section{Proof of Theorem \ref{thm:mse}}
Our proof follows the same steps as of   \citep[Theorem 2]{wang2017optimal}.
Without loss of generosity, we assume the context $x$ is a scalar. Let $\mathbb{X}_\tau:=\{x\in \mathcal{X}: D_\mathrm{KL}(x)<\tau\}$. The mean squared error can be decomposed into squared bias and variance,
\begin{align}
\label{eq:thm2-pf-step1}
    \text{MSE}(\widehat{V}_\mathrm{DR-IC}) = (\mathbb{E}[\widehat{V}_\mathrm{DR-IC}]-V^\pi)^2 + \text{Var}[\widehat{V}_\mathrm{DR-IC}].
\end{align}
For the bias part, notice we only need to consider for $x\in \mathbb{X}_\tau^c$, so
\begin{align}
    \mathbb{E}[\widehat{V}_\mathrm{DR-IC}]-V^\pi & = \mathbb{E}\Bigg[\sum_{a\in \mathcal{A}}\widehat{r}_{\mathrm{IB}}(x,a)\pi(a|x)\indic(x\in\mathbb{X}_\tau^c) \Bigg] - \mathbb{E}\Bigg[\sum_{a\in \mathcal{A}}\mathbb{E}[r|x,a]\pi(a|x)\indic(x\in\mathbb{X}_\tau^c) \Bigg] \nonumber \\
    & = \mathbb{E}_{\pi}\Bigg[\Big(\widehat{r}_{\mathrm{IB}}(x,a)-\mathbb{E}[r|x,a]\Big)\indic(x\in\mathbb{X}_\tau^c) \Bigg] \nonumber \\
    \label{eq:thm2-pf-step2}
    & = \mathbb{E}_{\pi}\Big[\epsilon_{\mathrm{IB}}(x,a)\indic(x\in\mathbb{X}_\tau^c) \Big].
\end{align}
Next, we derive the upper bound of the variance term in the MSE. For any random variable $X$ and $Y$,
$$
    \text{Var}(X+Y)\leq 2\text{Var}(X) + 2\text{Var}(Y).
$$
Thus,
\begin{align}
    \text{Var}[\widehat{V}_\mathrm{DR-IC}] 
    & = \text{Var}\Bigg[\frac{1}{n}\sum_{i=1}^n\sum_{a\in \mathcal{A}}\widehat{r}_{\mathrm{IB}}(x_i,a)\pi(a|x_i) + \frac{1}{n}\sum_{i=1}^n (r_i-\widehat{r}_{\mathrm{IB}}(x_i,a_i))w(x_i,a_i)\indic(x_i\in \mathbb{X}_\tau)\Bigg] \nonumber\\
    & \leq \frac{2}{n}\text{Var}\Bigg[\sum_{a\in \mathcal{A}}\widehat{r}_{\mathrm{IB}}(x,a)\pi(a|x)\Bigg] + \frac{2}{n}\text{Var}_{\mu}\Bigg[(r-\widehat{r}_{\mathrm{IB}}(x,a))w(x,a)\indic(x\in\mathbb{X}_\tau) \Bigg] \nonumber \\
    & \leq \frac{2}{n}\mathbb{E}\Bigg[\sum_{a\in \mathcal{A}}\widehat{r}_{\mathrm{IB}}(x,a)\pi(a|x)\Bigg]^2
    + \frac{2}{n}\mathbb{E}_{\mu}\text{Var}\Bigg[ (r-\widehat{r}_{\mathrm{IB}}(x,a))w(x,a)\indic(x\in \mathbb{X}_\tau)\Bigg| x,a \Bigg] \nonumber \\
    & \phantom{\leq \frac{2}{n}\mathbb{E}\Bigg[\sum_{a\in A}\widehat{r}_\text{ll}(a,x)\pi(a|x)\Bigg]^2+} 
    + \frac{2}{n}\text{Var}_{\mu}\mathbb{E}\Bigg[ (r-\widehat{r}_{\mathrm{IB}}(x,a))w(x,a)\indic(x\in \mathbb{X}_\tau)\Bigg| x,a  \Bigg] \nonumber \\
    & \leq \frac{2}{n}\mathbb{E}\Bigg[|\mathcal{A}|\sum_{a\in \mathcal{A}}\widehat{r}^2_{\mathrm{IB}}(x,a)\pi^2(a|x)\Bigg] + \frac{2}{n}\mathbb{E}_{\mu}\Bigg[\mathbb{E}(r-\widehat{r}_{\mathrm{IB}}(x,a))^2w^2(x,a) \Bigg] \nonumber \\
    & \phantom{\leq \frac{2}{n}\mathbb{E}\Bigg[|\mathcal{A}|\sum_{a\in \mathcal{A}}\widehat{r}^2_\text{ll}(a,x)\pi^2(a|x)\Bigg] +} 
    + \frac{2}{n} \mathbb{E}_{\mu} \mathbb{E}\Bigg[ (r-\widehat{r}_{\mathrm{IB}}(x,a))w(x,a)\indic(x\in \mathbb{X}_\tau)\Bigg| x,a  \Bigg]^2 \nonumber \\
    & \leq \frac{2}{n} |\mathcal{A}|\,\mathbb{E}\Bigg[ \sum_{a\in \mathcal{A}}\widehat{r}^2_{\mathrm{IB}}(x,a)\pi(a|x) \Bigg] + \frac{2}{n} \mathbb{E}_{\mu}\Bigg[w_{\max}^2(\widehat{r}_{\mathrm{IB}}(x,a)-\mathbb{E}[r|x,a])^2 \Bigg] \nonumber \\
    & \phantom{\leq \frac{2}{n} |\mathcal{A}|\,\mathbb{E}\Bigg[ \sum_{a\in A}\widehat{r}^2_\text{ll}(a,x)\pi(a|x) \Bigg] + } 
    + \frac{2}{n}\mathbb{E}_{\mu}\mathbb{E}\Bigg[ (r-\widehat{r}_{\mathrm{IB}}(x,a))^2w^2(x,a)\Bigg|x,a \Bigg] \nonumber \\
    \label{eq:thm2-pf-step3}
    & \leq \frac{2}{n}|\mathcal{A}|\,\mathbb{E}_\pi[R^2_{\max}] + \frac{2}{n} \mathbb{E}_{\mu}\Big\{w^2_{\max}\epsilon^2_{\mathrm{IB}}(x,a)+ \mathbb{E}[(r-\widehat{r}_{\mathrm{IB}}(x,a))^2\mid x,a]\Big\}.
\end{align}
Using \eqref{eq:thm2-pf-step2} and \eqref{eq:thm2-pf-step3} in \eqref{eq:thm2-pf-step1}, we get the desired bound.

\section{Additional Experiments and Implementation  Details}

We use $10$ multi-class classification datasets from the \texttt{UCI Machine Learning Repository} (available at \url{https://archive.ics.uci.edu/ml/index.php}) \citep{dua2019uci}. We use these datasets in the comma-separated values format provided by the \texttt{Datahub Machine Learning Repository} (available at \url{https://datahub.io/machine-learning}). Information of the datasets are given in Table \ref{tab:datasets}.

\begin{table}[h!]
\begin{center}
{\scriptsize
\begin{tabular}{|c|c|c|c|c|c|c|c|c|c|c|}
\hline
Dataset & Glass & Ecoli & Wdbc & Vehicle & Yeast & Page-Blocks & OptDigits & SatImage & PenDigits & Letter\\
\hline\hline
Contexts dimension & 9 & 7 & 30 & 18 & 8 & 10 & 64 & 36 & 16 & 16\\
\hline
No. of actions & 6 & 8 & 2 & 4 & 10 & 5 & 10 & 6 & 10 & 26\\
\hline
Max sample size & 214 & 336 & 569 & 846 & 1484 & 5473 & 5620 & 6435 & 10992 & 20000\\
\hline
\end{tabular}}
\caption{Dataset information.}
\label{tab:datasets}
\end{center}
\end{table}

We now describe the hyperparameter grids we use in our experiments. We adapt from \citep{dudik2014doubly,wang2017optimal,su2020doubly} for these grids. \\
(1) For our \textbf{DR-IC estimator}, we have two hyperparameters, namely, bandwidth $h_n$'s and the switching $\tau$. With $\tau=\infty$, we first find the best $h_n$ from a grid of 30 geometrically spaced values between the $0.01$ and $15.0$. With these best bandwidths for different $n$'s, we choose the switching parameter from a grid of 30 geometrically spaced values between the $0.01$ quantile and $1.0$ quantile of the KL divergence of the contexts observed from the data. We also include the $0^{\mathrm{th}}$ quantile in this grid for the switching parameter. \\
(2) For the \textbf{DR-OS estimator}, we choose the shrinkage coefficients from a grid of 30 geometrically spaced values between $0.01\times(w_{0.05})^2$ and $100\times(w_{0.95})^2$ where $w_{0.05}$ and $w_{0.95}$ are the 0.05 and 0.95 quantile of the importance weights ($\pi/\mu$) observed in the data. \\
(3) For the \textbf{Switch-DR estimator}, we choose the switching parameter from a grid of 25 exponentially spaced values between the $0.05$ quantile and $0.95$ quantile of the importance weights observed in the data. \\
We note that we use 15 grid values for the datasets with maximum sample size larger than 5000 due to the hardware limitations.

We use the \texttt{Open Bandit Pipeline} (available at \url{https://github.com/st-tech/zr-obp}) library to implement our DR-IC estimator and benchmark it with the other estimators. We worked with this commit-version (\url{https://github.com/st-tech/zr-obp/commit/a4b61e9e14d1954aa2953d2b21b4e710a8a725d8}) of the \texttt{Open Bandit Pipeline} library and provide the same in our supplementary material. We now describe practical implementation details of our DR-IC estimator, focusing on the discrete action setting.

Recall from \eqref{eq:r-IB-1} that
\begin{align*} 
 & \widehat{\mathbf{r}}_{\mathrm{IB}}(\widetilde{Z}) =\widetilde{Z}\widehat{\theta}_{\mathrm{ls}} + \Sigma(\widetilde{\mathbf{r}}, \mathbf{r})\Sigma_\mathbf{r}^{-1}\left[\mathrm{diag}\{\Sigma(\widetilde{\mathbf{r}}, \mathbf{r})\Sigma_\mathbf{r}^{-1}\bm{1}_n\}\right]^{-1}(\mathbf{r} - Z\widehat{\theta}_{\mathrm{ls}}).
\end{align*}
We use the Gaussian kernel for $\Sigma(\widetilde{\mathbf{r}}, \mathbf{r})$ as in \eqref{eq:gaussian-covf}. We use the traditional logistic regression as the estimate $\widehat{\theta}_{\mathrm{ls}}$ with three-fold cross fitting. Finally, we set $\hat{\Sigma}_\mathbf{r} = \mathrm{Var}(\mathbf{r} - Z\widehat{\theta}_{\mathrm{ls}}) I$ where $\mathrm{Var}(x)$ is the variance of vector $x$ and $I$ is an appropriate sized identity matrix. Since $\Sigma_\mathbf{r}$ is unknown in practice, we use the above estimate under an independent homoscedastic assumption.

\begin{figure*}[h!]
	\centering
	\begin{minipage}{.24\textwidth}
		\centering
		\includegraphics[width=\linewidth]{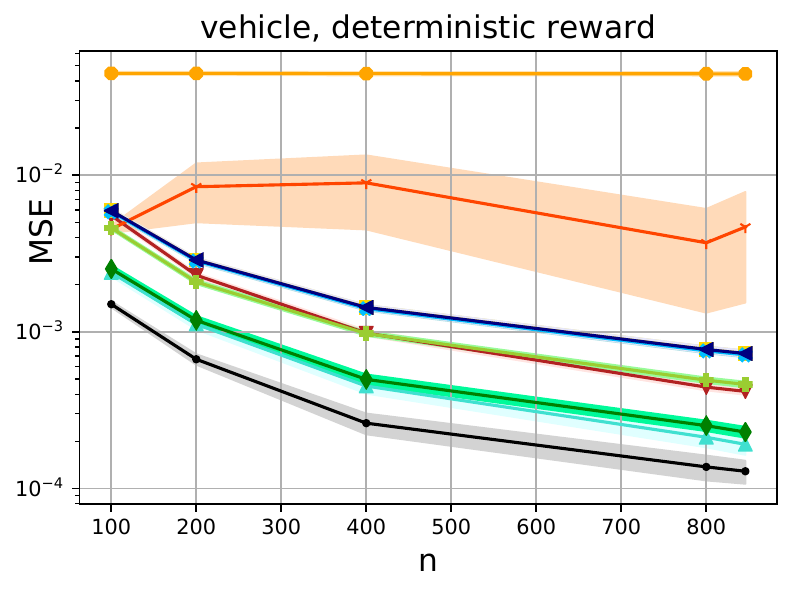}
	\end{minipage}
	\begin{minipage}{.24\textwidth}
		\centering
		\includegraphics[width=\linewidth]{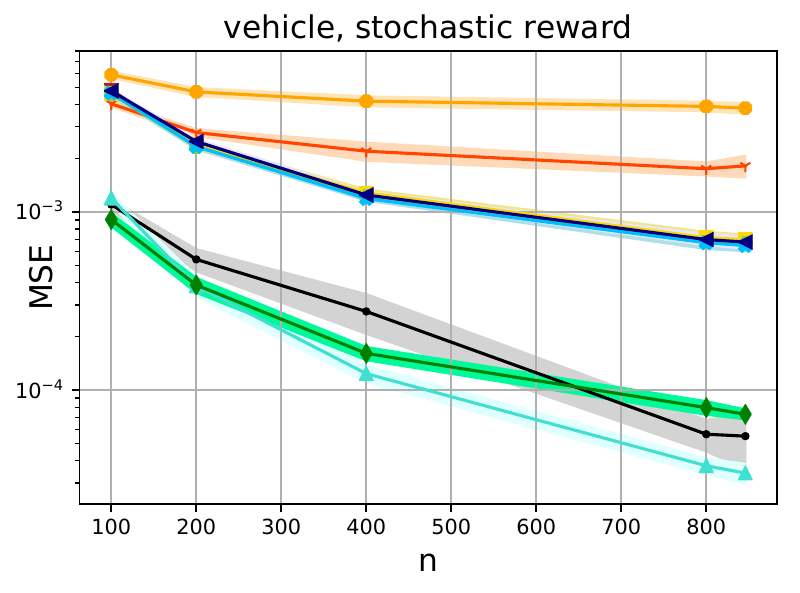}
	\end{minipage}
	\begin{minipage}{.24\textwidth}
		\centering
		\includegraphics[width=\linewidth]{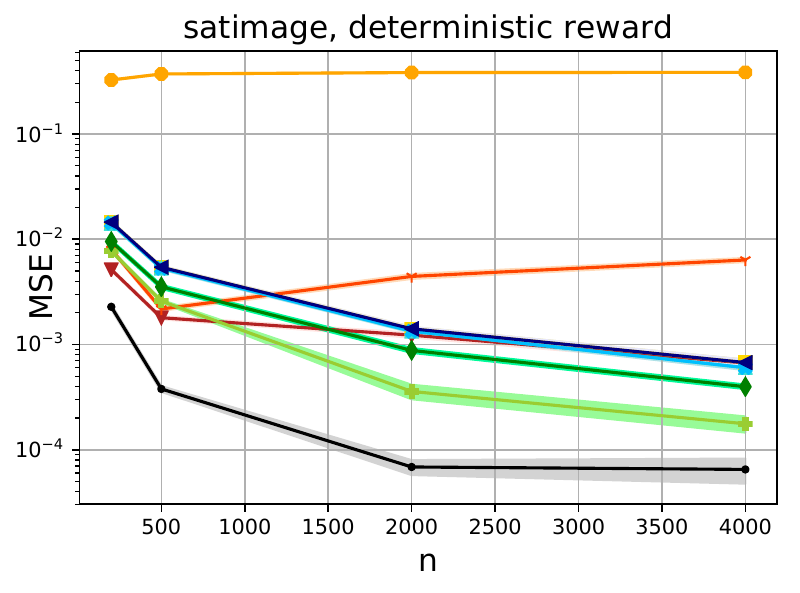}
	\end{minipage}
	\begin{minipage}{.24\textwidth}
		\centering
		\includegraphics[width=\linewidth]{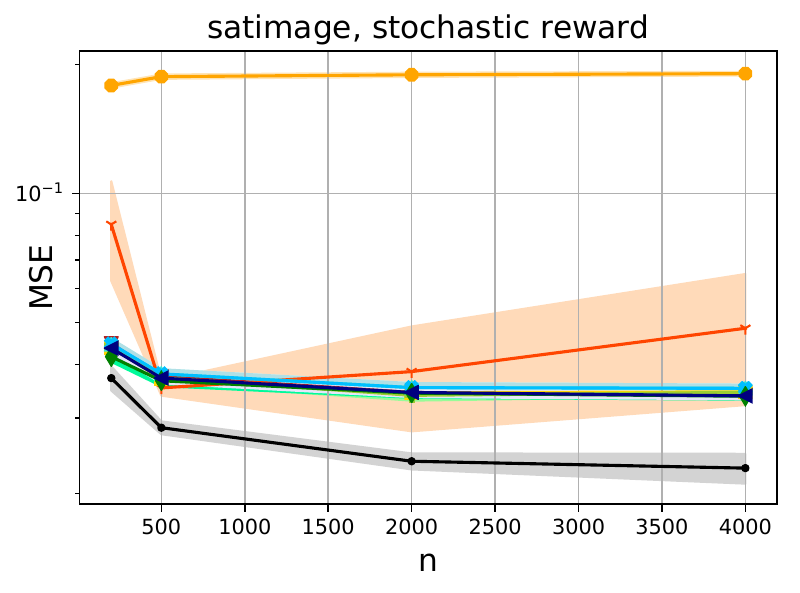}
	\end{minipage}
	\centering
	\begin{minipage}{.24\textwidth}
		\centering
		\includegraphics[width=\linewidth]{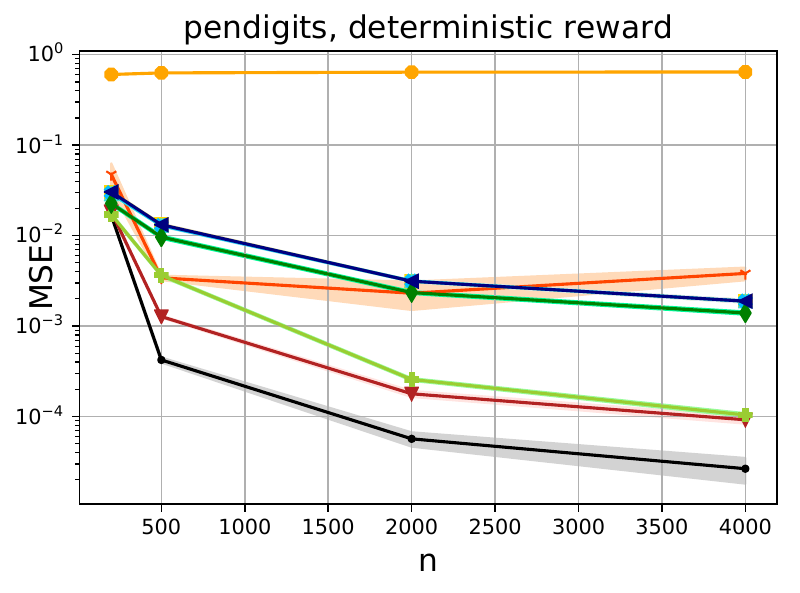}
	\end{minipage}
	\begin{minipage}{.24\textwidth}
		\centering
		\includegraphics[width=\linewidth]{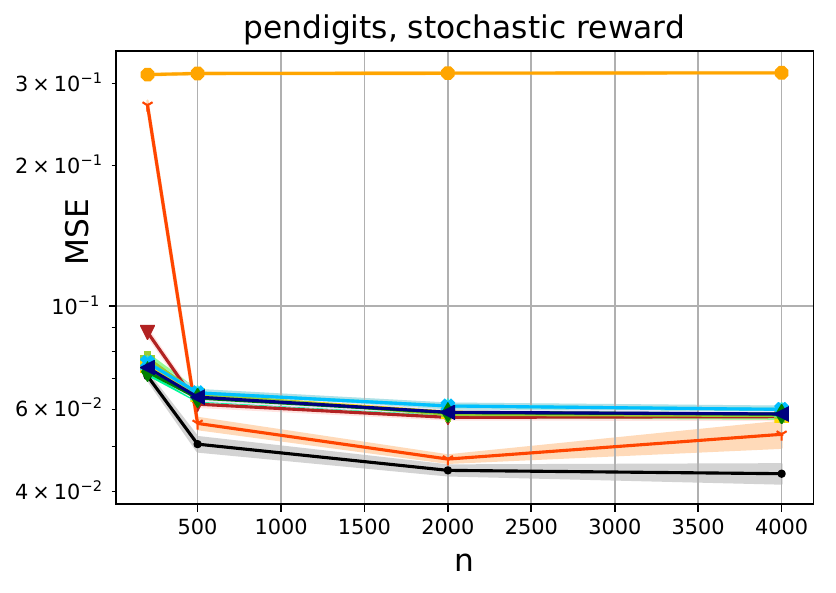}
	\end{minipage}
	\begin{minipage}{.24\textwidth}
		\centering
		\includegraphics[width=\linewidth]{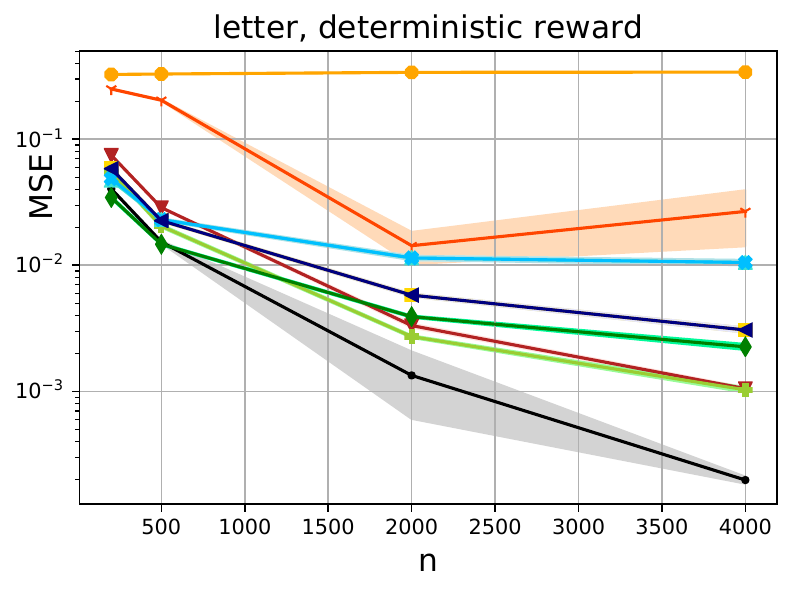}
	\end{minipage}
	\begin{minipage}{.24\textwidth}
		\centering
		\includegraphics[width=\linewidth]{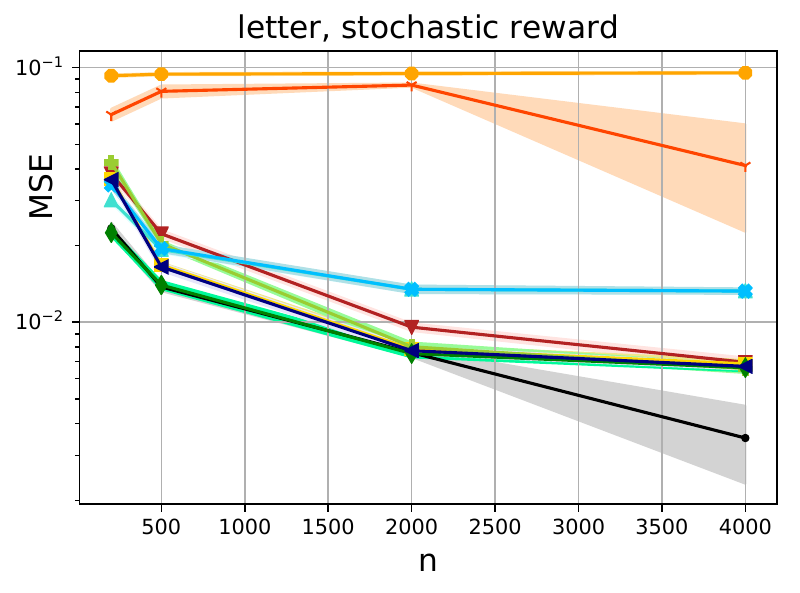}
	\end{minipage}
	
	\captionof{figure}{\textit{Off-policy evaluation simulation results.} MSE versus sample sizes $n$ for two different reward distributions are showcased for all 10 UCI datasets described in Table \ref{tab:datasets} }
	\label{figcomp-appendix}
\end{figure*}

We now summarize our results with the plots for all UCI datasets presented in Fig. \ref{figcomp-appendix}. At foremost, we present the superior performance statistics of our estimator compared to seven different estimators from the off-policy evaluation literature in Tables \ref{tab:deterministic-alg-comp} and \ref{tab:stochastic-alg-comp}. The uniformly superior performance of the information-borrowed DM (DR-IC $(\tau=0)$) over the traditional DM corroborates the recent findings of \cite{su2020doubly} that the choice of the DM is indeed important. We also demonstrate the uniformly dominating performance of the proposed DR-IC method over its counterparts across all the datasets, especially for the deterministic reward model. Even for the stochastic reward model, DR-IC beats its counterparts in most cases.   Note that, for the current experiments, the stochasticity in the reward model causes the rewards to have a discrete jump (between $0$ and $1$) due to the model choice. We particularly expect a stronger performance for the proposed approach when the considered reward model is continuous in context and action space because the DR-IC approach can borrow information effectively from similar contexts. This suggests that the proposed approach is appropriate for a wide range of datasets and reward models. Furthermore, the KL approach to switching dominates the importance weights based switching in $12$ cases, and is equivalent in $7$ cases out of the $20$ cases considered. This clearly suggests that the KL approach is a  preferable and robust choice of switching across a variety of datasets.

\begin{table}[h!]
\begin{center}
{\small
\begin{tabular}{|c|c|c|c|c|c|c|c|}
\hline
Alg. & DM & DR & MRDR & DR-OS (oracle) & DR-OS & Switch-DR (oracle) & Switch-DR\\
\hline\hline
DR-IC (oracle) & 10/0/0 & 10/0/0 & 10/0/0 & 10/0/0 & 10/0/0 & 10/0/0 & 10/0/0 \\
\hline
DR-IC & 10/0/0 & 8/2/0 & 7/0/3 & 5/1/4 & 8/1/1 & 3/1/6 & 8/2/0\\
\hline
DR-IC ($\tau=0$) & 10/0/0 & - & - & - & - & - & -\\
\hline
\end{tabular}}
\caption{\textbf{Performance Statistics:} No. of wins/No. of draws/No. of losses by our estimator during off-policy evaluation training compared to others in the deterministic reward setting. Here the MSE metric is used for the comparison with no. of dataset as units. }
\label{tab:deterministic-alg-comp}
\end{center}
\end{table}

\begin{table}[h!]
\begin{center}
{\small
\begin{tabular}{|c|c|c|c|c|c|c|c|}
\hline
Alg. & DM & DR & MRDR & DR-OS (oracle) & DR-OS & Switch-DR (oracle) & Switch-DR\\
\hline\hline
DR-IC (oracle) & 10/0/0 & 10/0/0 & 10/0/0 & 7/1/2 & 10/0/0 & 6/1/3 & 10/0/0 \\
\hline
DR-IC & 10/0/0 & 3/6/1 & 3/6/1 & 2/5/3 & 5/3/2 & 4/2/4 & 4/5/1\\
\hline
DR-IC ($\tau=0$) & 9/1/0 & - & - & - & - & - & -\\
\hline
\end{tabular}}
\caption{\textbf{Performance Statistics:} No. of wins/No. of draws/No. of losses by our estimator during off-policy evaluation training compared to others in the stochastic reward setting. Here the MSE metric is used for the comparison with no. of dataset as units.}
\label{tab:stochastic-alg-comp}
\end{center}
\end{table}

\begin{figure*}[h!]
    \centering
	\includegraphics[width=\linewidth]{Figures/legend.pdf}
	\centering
	\begin{minipage}{.24\textwidth}
		\centering
		\includegraphics[width=\linewidth]{Figures/ecoli_stoc_reward_False.pdf}
	\end{minipage}
	\begin{minipage}{.24\textwidth}
		\centering
		\includegraphics[width=\linewidth]{Figures/ecoli_stoc_reward_True.pdf}
	\end{minipage}
	\begin{minipage}{.24\textwidth}
		\centering
		\includegraphics[width=\linewidth]{Figures/glass_stoc_reward_False.pdf}
	\end{minipage}
	\begin{minipage}{.24\textwidth}
		\centering
		\includegraphics[width=\linewidth]{Figures/glass_stoc_reward_True.pdf}
	\end{minipage}
	\centering
	\begin{minipage}{.24\textwidth}
		\centering
		\includegraphics[width=\linewidth]{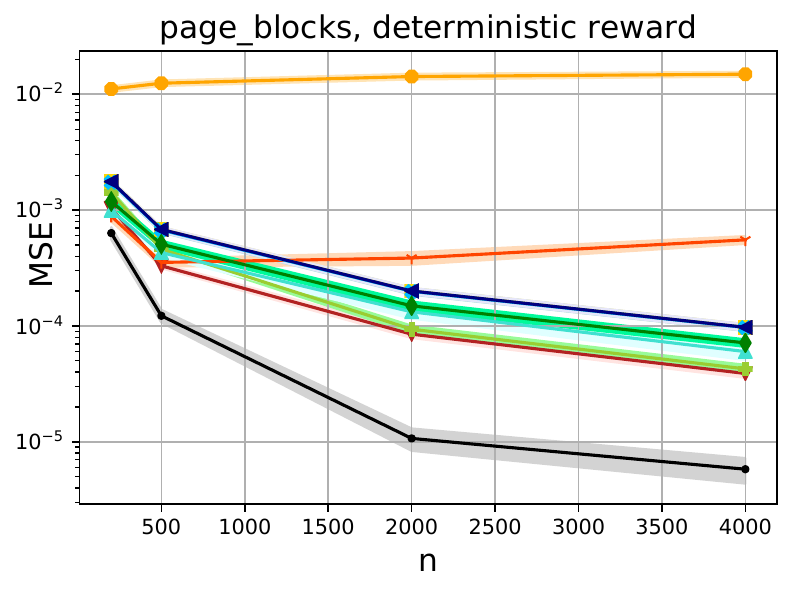}
	\end{minipage}
	\begin{minipage}{.24\textwidth}
		\centering
		\includegraphics[width=\linewidth]{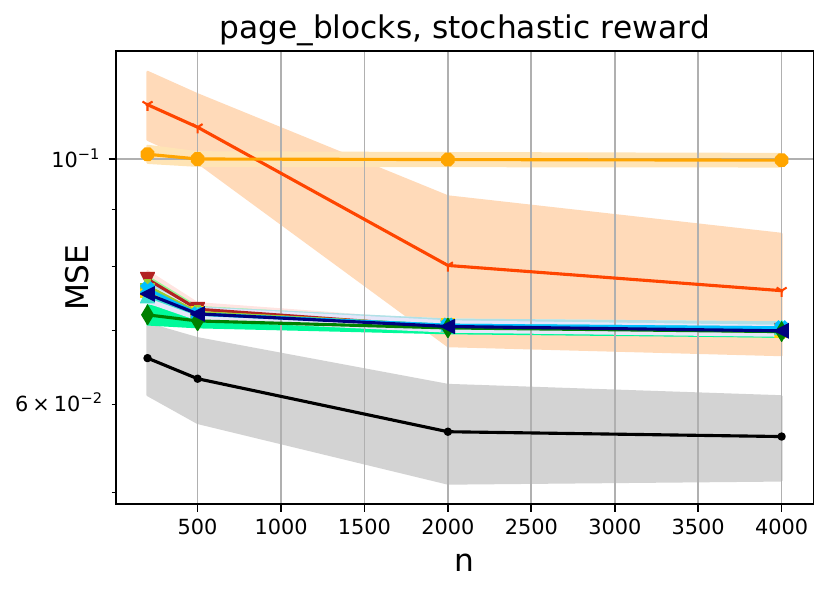}
	\end{minipage}
	\begin{minipage}{.24\textwidth}
		\centering
		\includegraphics[width=\linewidth]{Figures/yeast_stoc_reward_False.pdf}
	\end{minipage}
	\begin{minipage}{.24\textwidth}
		\centering
		\includegraphics[width=\linewidth]{Figures/yeast_stoc_reward_True.pdf}
	\end{minipage}
	\centering
	\begin{minipage}{.24\textwidth}
		\centering
		\includegraphics[width=\linewidth]{Figures/optdigits_stoc_reward_False.pdf}
	\end{minipage}
	\begin{minipage}{.24\textwidth}
		\centering
		\includegraphics[width=\linewidth]{Figures/optdigits_stoc_reward_True.pdf}
	\end{minipage}
	\begin{minipage}{.24\textwidth}
		\centering
		\includegraphics[width=\linewidth]{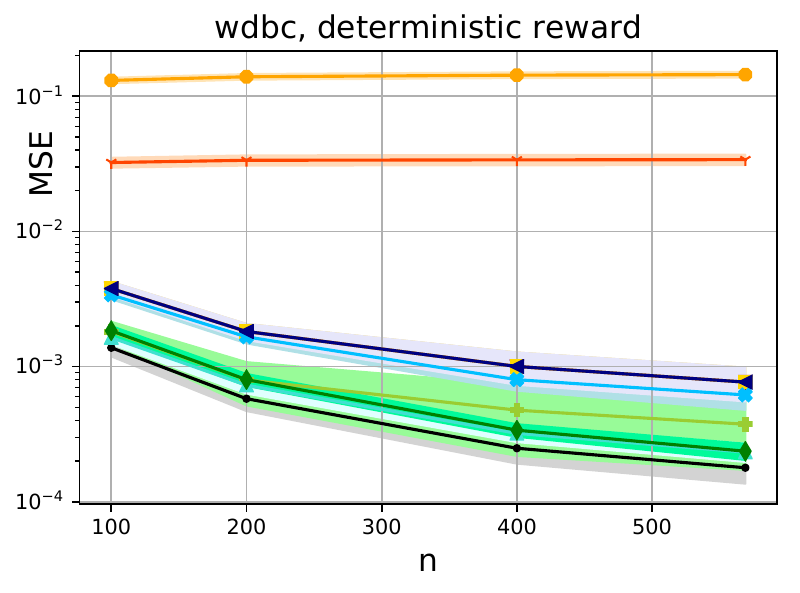}
	\end{minipage}
	\begin{minipage}{.24\textwidth}
		\centering
		\includegraphics[width=\linewidth]{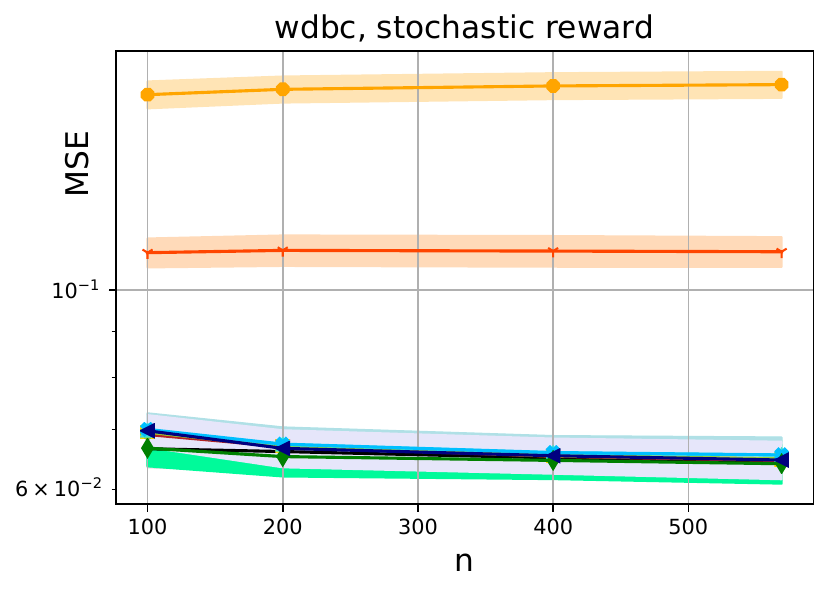}
	\end{minipage}
\end{figure*}

\section{Experiments with Nadaraya-Watson regression (non-parametric method)}

Our proposed estimator is a semiparametric estimator, with a kernel function to model the correlation structure. In this section, we compare the performance of our proposed estimators with a standard Nadaraya-Watson (NW) estimator. An NW estimator is a very well known non-parametric technique for kernel regression. Given the rewards  $r_i$ as responses, the context action pairs $z_i$ as predictors and a kernel function K, the NW estimate $\hat{r}_{\mathrm{NW}}$ at a point $z$ is given by
$$
\hat{r}_{\mathrm{NW}}(z)= \frac{\sum_{i=1}^n r_i K\left((z - z_i)/h\right)}{\sum_{j=1}^n K\left((z - z_j)/h\right)}
$$
Here, $h$ is a bandwidth parameter, which is often optimized (using cross validation, for example) given the data and acts as a hyperparameter. In this section we compare our estimators with the DM and DR estimators that use NW kernel regression models for estimating the rewards. We use Sklearn's (\url{https://scikit-learn.org/stable/related_projects.html}) implementation of NW kernel regression for our experiments and this is made available in our code. In these experiments, we use radial basis functions as the kernel, that is, $K((z - z_i)/h) = \exp(-h\| z-z_i \|_2)$. For the hyperparameter $h$, we choose the best among the grid of 20 logarithmically spaced values between 0.01 and 100 via the one-leave-out cross validation technique.

\begin{figure*}[ht]
	\centering
	\begin{minipage}{.24\textwidth}
		\centering
		\includegraphics[width=\linewidth]{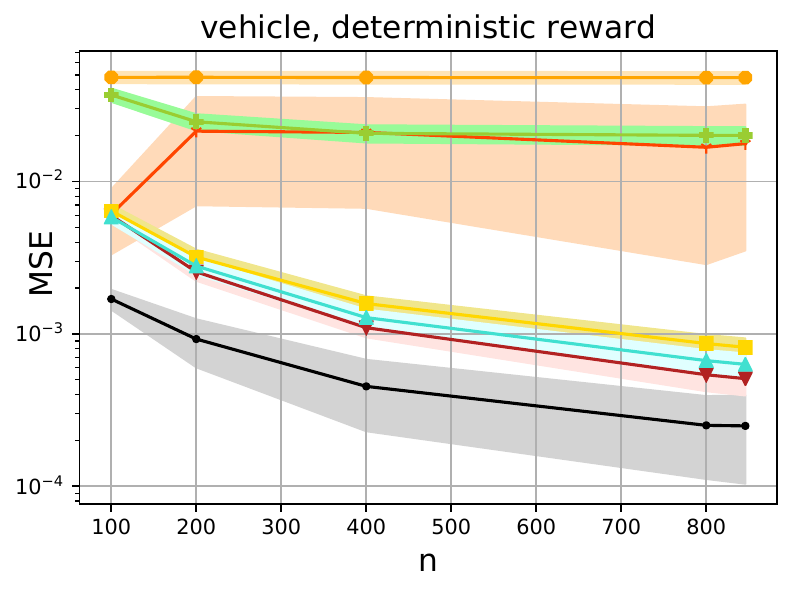}
	\end{minipage}
	\begin{minipage}{.24\textwidth}
		\centering
		\includegraphics[width=\linewidth]{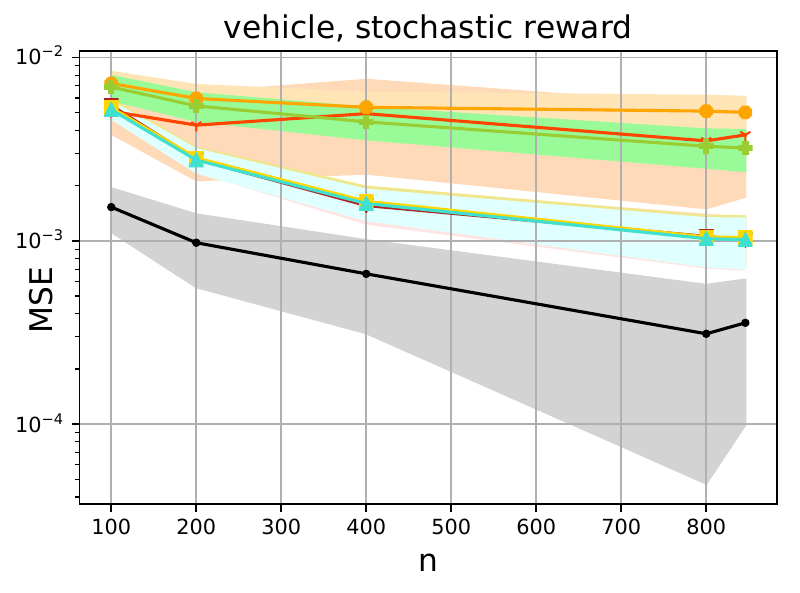}
	\end{minipage}
	\captionof{figure}{NW based DM and DR comparisons.}
	\label{figcomp-nw-1}
\end{figure*}

The experiments show the superior performance of our approach in most cases, as given in  Fig.\ref{figcomp-nw-1}. Interestingly, both our approach and NW approach  have a single bandwidth parameter. However, our  DM-IB (dr-ic( $\tau=0$)) method has the importance weight terms in the kernel (Eq. $8$) which implicitly allows an adaptive   bandwidth for different actions. This allows more flexible  borrowing, which is not possible in a traditional implementation of NW estimator. This is crucial because the information shared across $(x_1, a_1)$ and $(x_2, a_1)$ should be different from information shared across $(x_1, a_2)$ and $(x_2,a_2)$ in general, when $a_1 \neq a_2.$ \\
Our formulation allows  borrowing of information across $x_1$ and $x_2$ {\it only} when the corresponding actions are same, through an indicator function. A traditional nonparametric estimator will borrow information even if the actions are different, leading to higher bias, especially when the actions are nominal. Traditional nonparametric methods also suffer because their non-asymptotic terms become very large, especially in higher dimensional problems, as pointed out in \cite{wang2017optimal} and the references therein.

\begin{figure*}[ht]
    \centering
	\includegraphics[width=\linewidth]{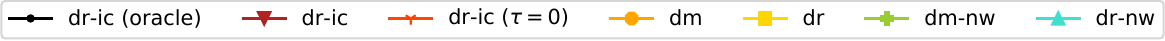}
	\centering
	\begin{minipage}{.24\textwidth}
		\centering
		\includegraphics[width=\linewidth]{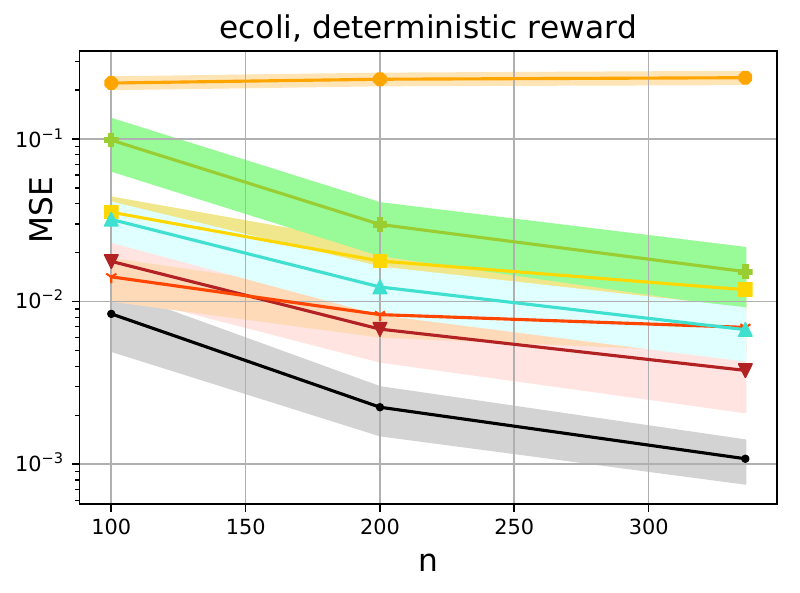}
	\end{minipage}
	\begin{minipage}{.24\textwidth}
		\centering
		\includegraphics[width=\linewidth]{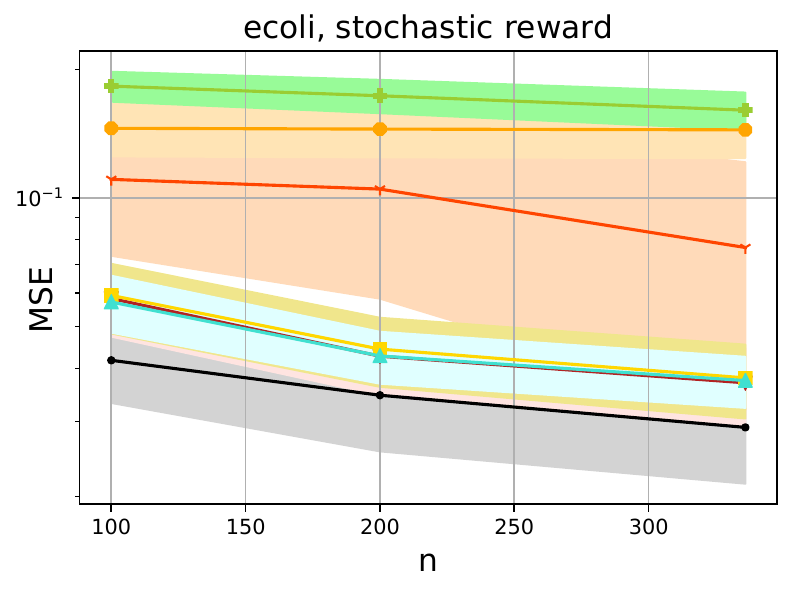}
	\end{minipage}
	\begin{minipage}{.24\textwidth}
		\centering
		\includegraphics[width=\linewidth]{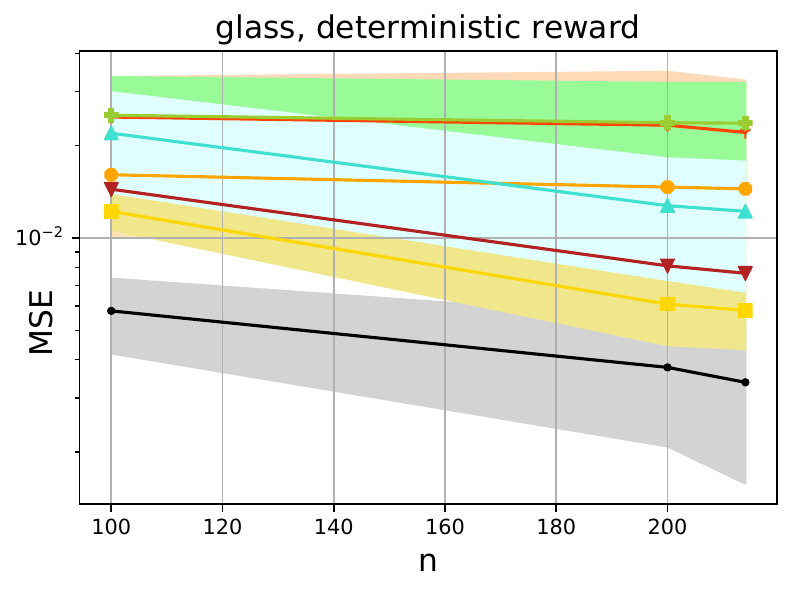}
	\end{minipage}
	\begin{minipage}{.24\textwidth}
		\centering
		\includegraphics[width=\linewidth]{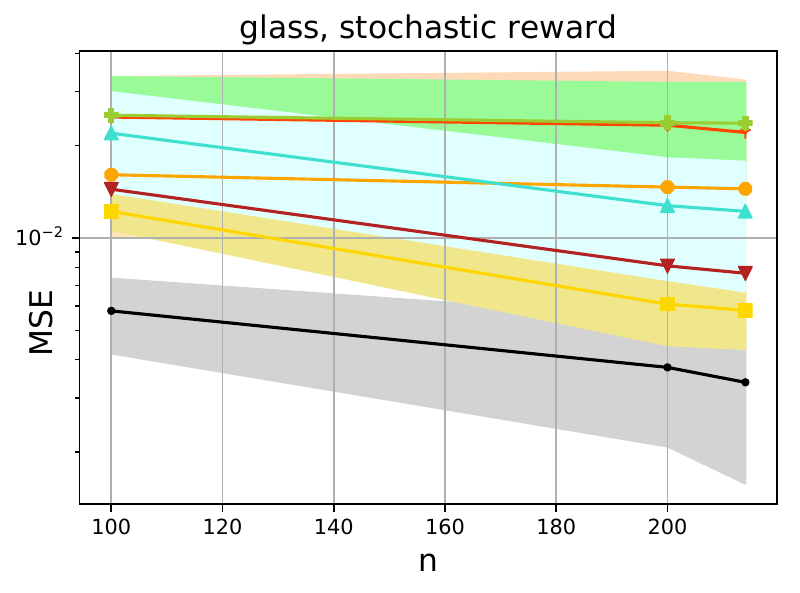}
	\end{minipage}
\end{figure*}
\begin{figure*}[ht]
	\centering
	\begin{minipage}{.24\textwidth}
		\centering
		\includegraphics[width=\linewidth]{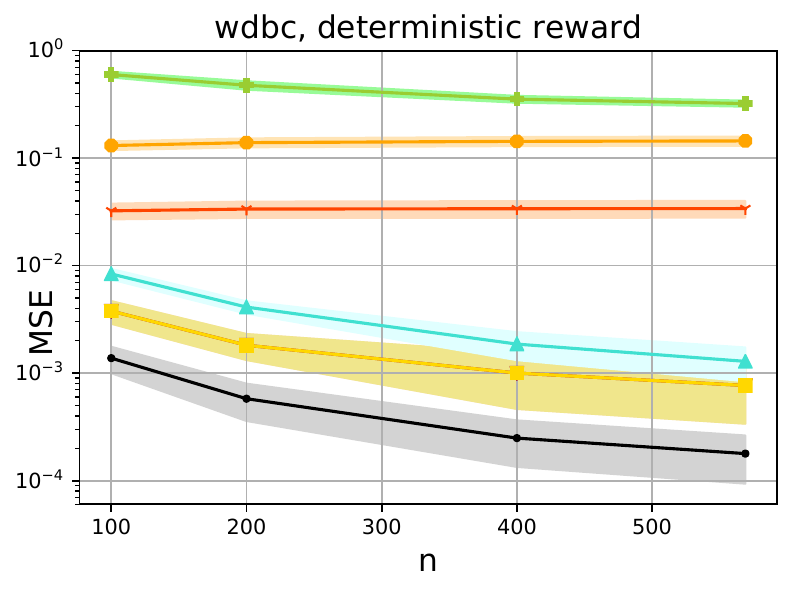}
	\end{minipage}
	\begin{minipage}{.24\textwidth}
		\centering
		\includegraphics[width=\linewidth]{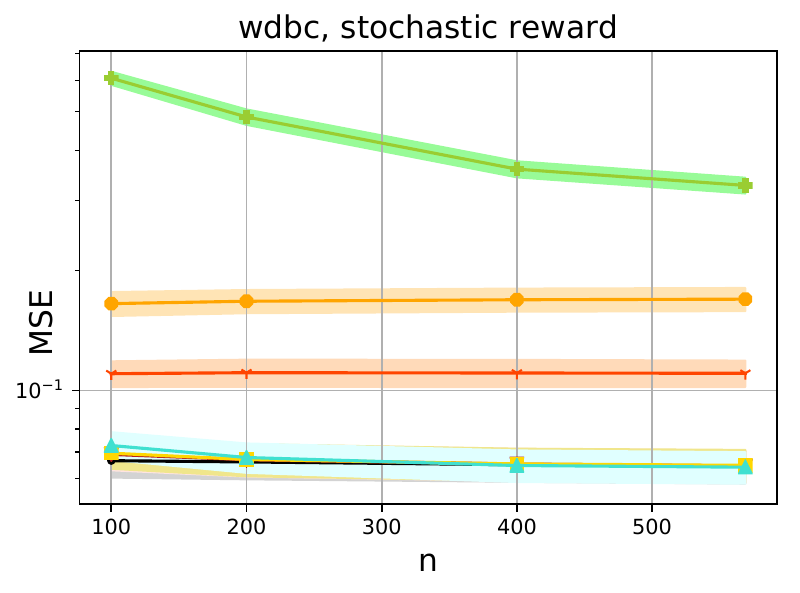}
	\end{minipage}
	\begin{minipage}{.24\textwidth}
		\centering
		\includegraphics[width=\linewidth]{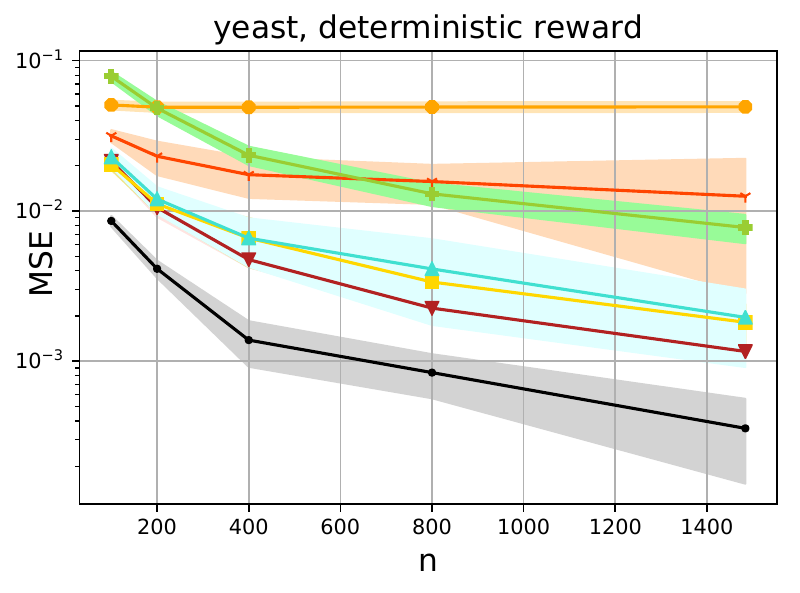}
	\end{minipage}
	\begin{minipage}{.24\textwidth}
		\centering
		\includegraphics[width=\linewidth]{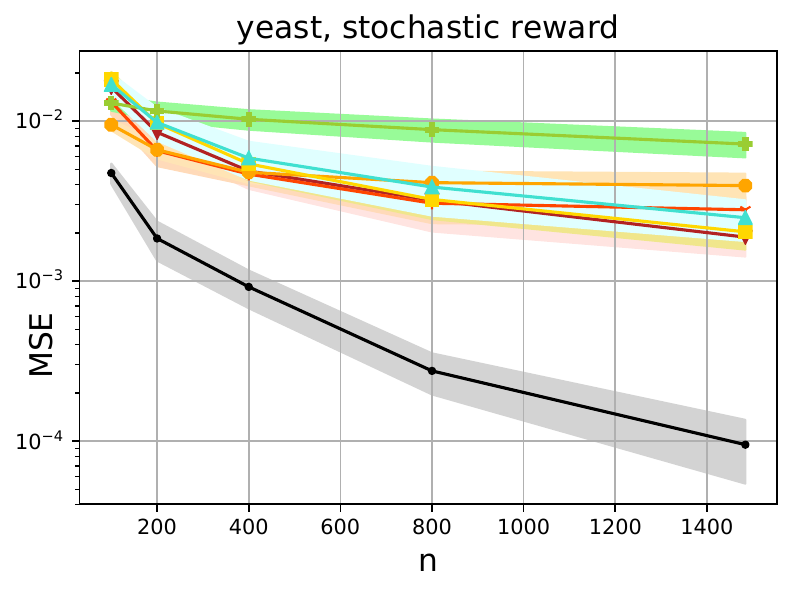}
	\end{minipage}
\end{figure*}

\begin{figure*}[ht]
	\centering
	\begin{minipage}{.24\textwidth}
		\centering
		\includegraphics[width=\linewidth]{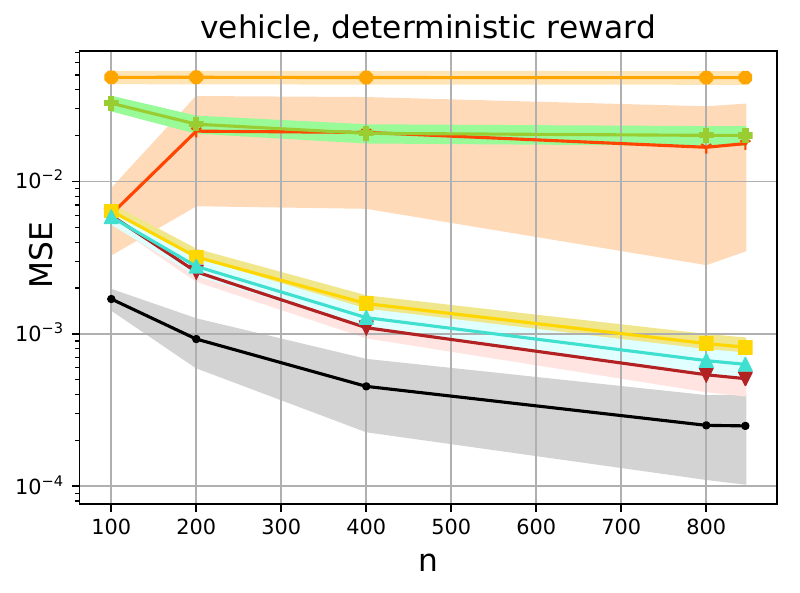}
	\end{minipage}
	\begin{minipage}{.24\textwidth}
		\centering
		\includegraphics[width=\linewidth]{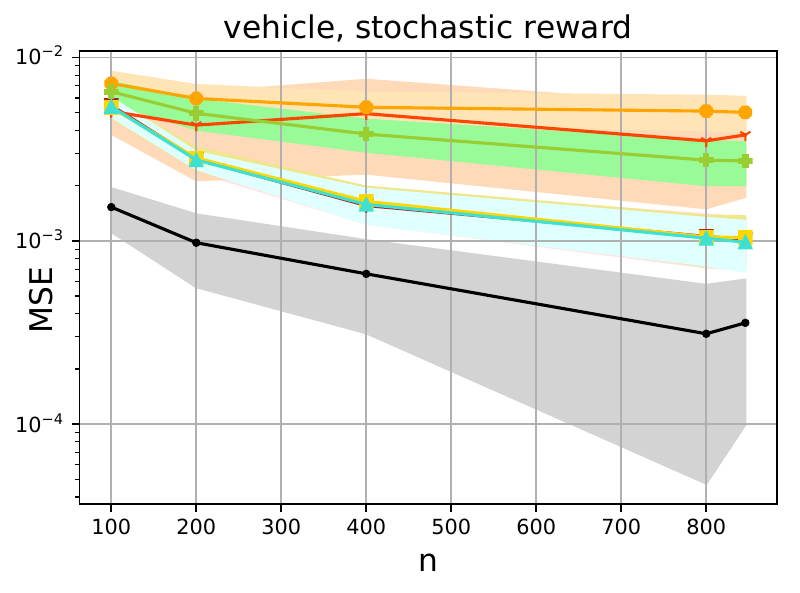}
	\end{minipage}
	\captionof{figure}{NW based DM and DR comparisons with indicator approximation for action covariates.}
	\label{figcomp-nw-2}
\end{figure*}

For a more fair comparison, we also emulate the indicator function  for action covariates by choosing appropriate fixed bandwidth. These experiments, whose results are given in  Fig.\ref{figcomp-nw-2}, still show superior performance of our approach in most cases. The reason is that the role of the kernel function is fundamentally different in the two approaches. In the proposed estimator, the parametric estimate through the ridge regression performs the overall borrowing of information, while the bandwidth parameter controls the differential borrowing of information across context-action pairs when the actions are same (in an adaptive manner depending on the importance weights), which leads to a significant improvement in practical performance. As an example, if the bandwidth parameter is very small, resulting in no borrowing of information, the estimate for that context action pair reduces to the ridge regression estimate. In contrast, the NW estimator has a bandwidth parameter which completely controls the  borrowing information across the observations. Thus, if the bandwidth parameter is very small, the final estimate reduces to  Dirac delta spikes with no borrowing among observations whatsoever.

\begin{figure*}[ht]
    \centering
	\includegraphics[width=\linewidth]{Figures/legend_nw.pdf}
	\centering
	\begin{minipage}{.24\textwidth}
		\centering
		\includegraphics[width=\linewidth]{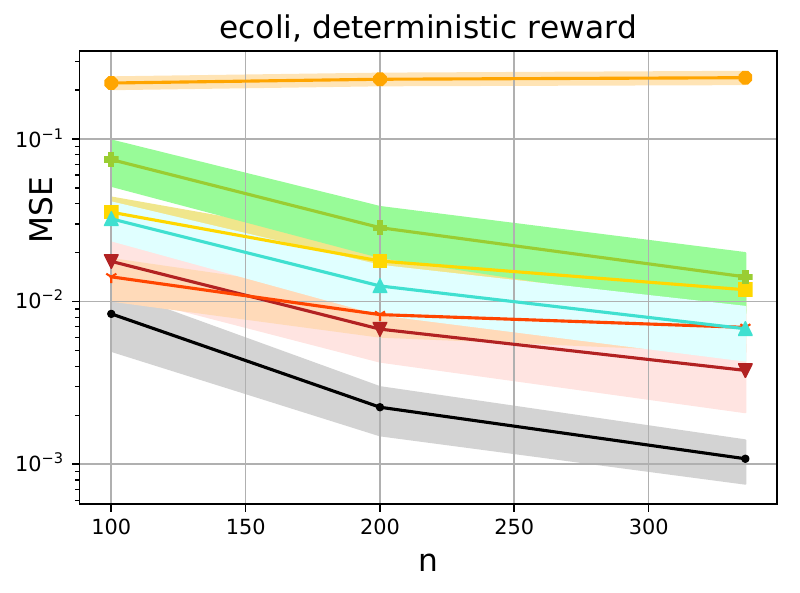}
	\end{minipage}
	\begin{minipage}{.24\textwidth}
		\centering
		\includegraphics[width=\linewidth]{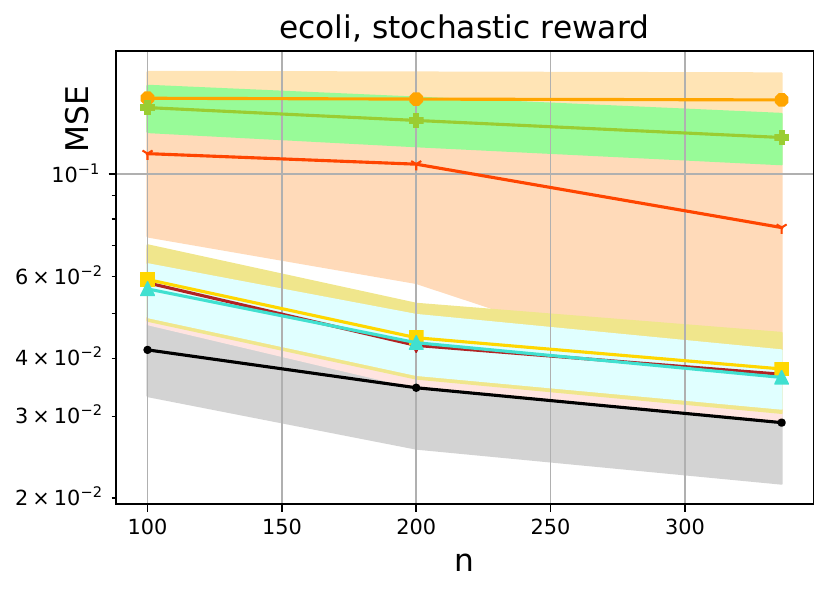}
	\end{minipage}
	\begin{minipage}{.24\textwidth}
		\centering
		\includegraphics[width=\linewidth]{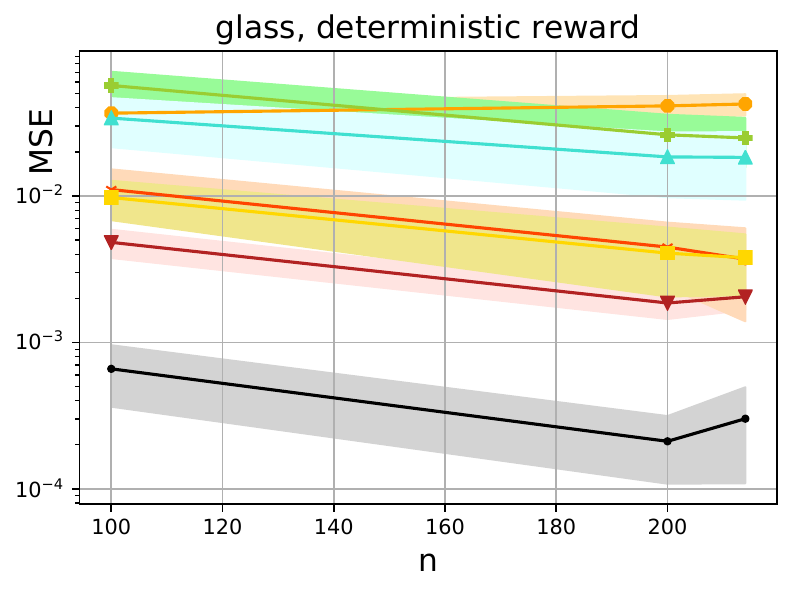}
	\end{minipage}
	\begin{minipage}{.24\textwidth}
		\centering
		\includegraphics[width=\linewidth]{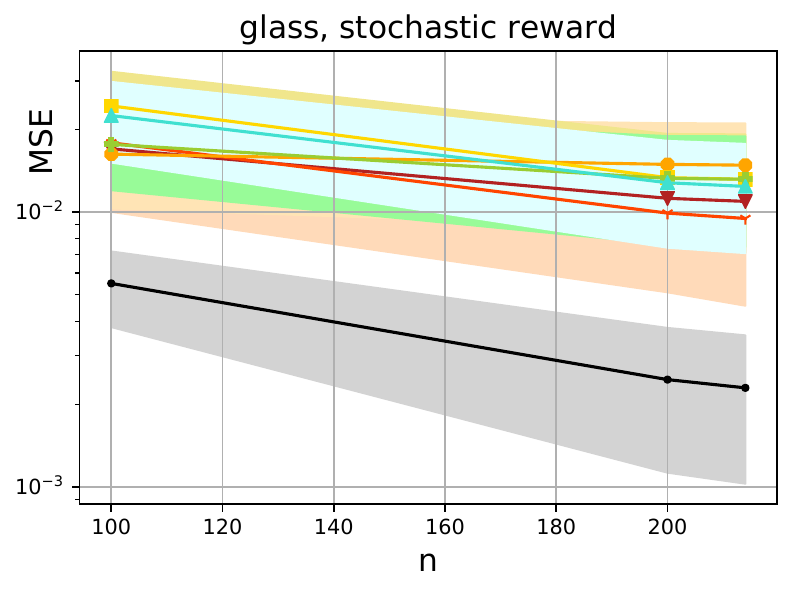}
	\end{minipage}
	\centering
	\begin{minipage}{.24\textwidth}
		\centering
		\includegraphics[width=\linewidth]{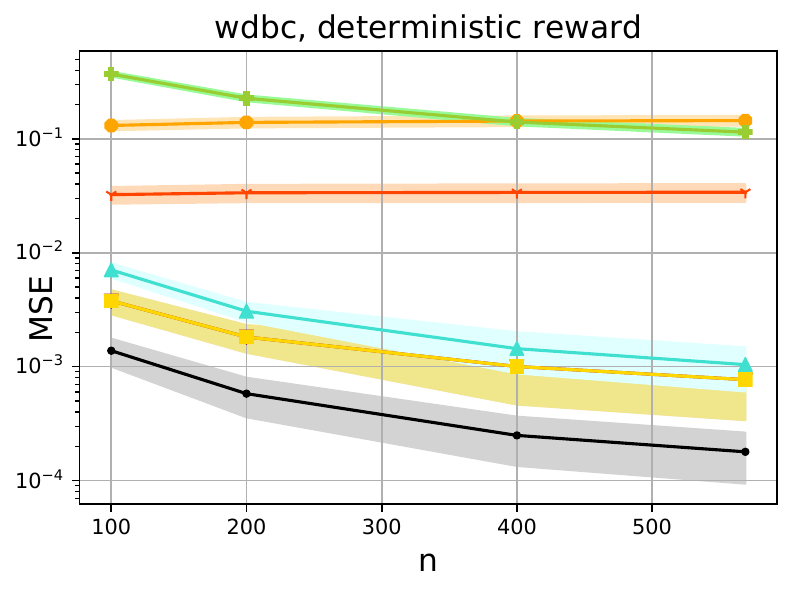}
	\end{minipage}
	\begin{minipage}{.24\textwidth}
		\centering
		\includegraphics[width=\linewidth]{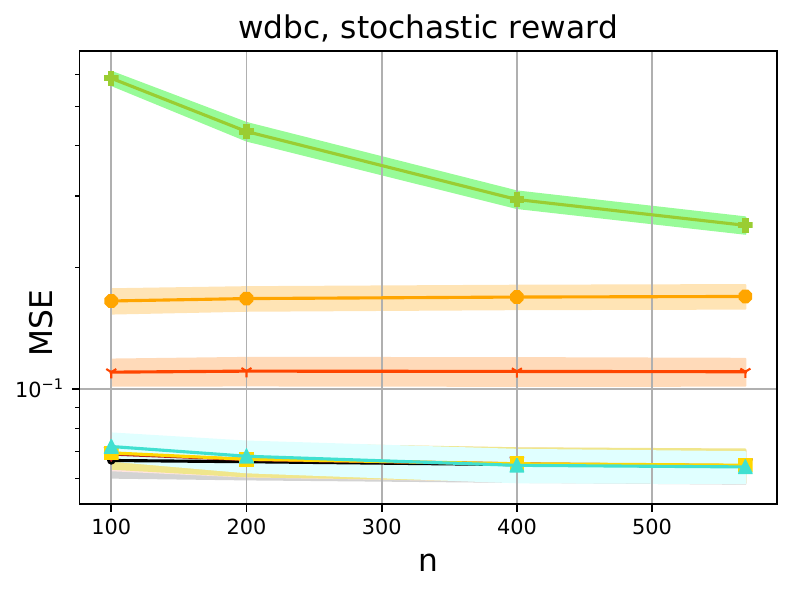}
	\end{minipage}
	\begin{minipage}{.24\textwidth}
		\centering
		\includegraphics[width=\linewidth]{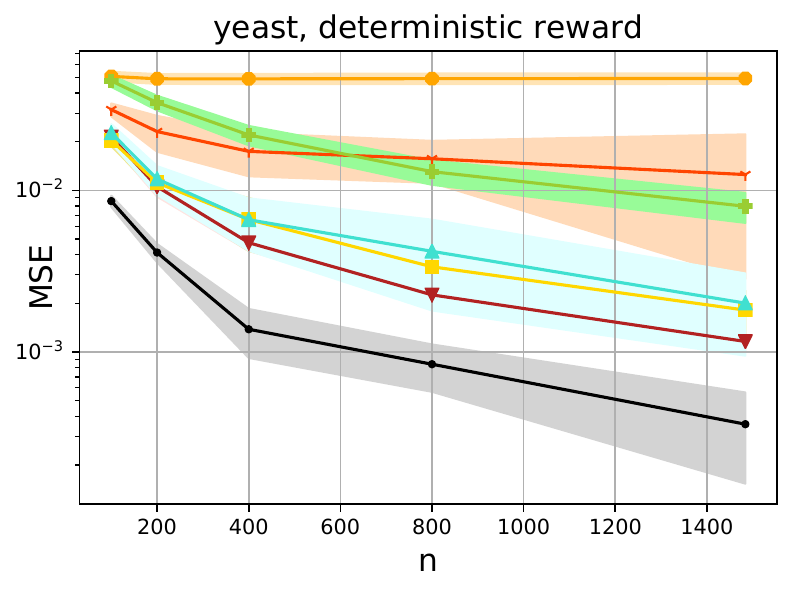}
	\end{minipage}
	\begin{minipage}{.24\textwidth}
		\centering
		\includegraphics[width=\linewidth]{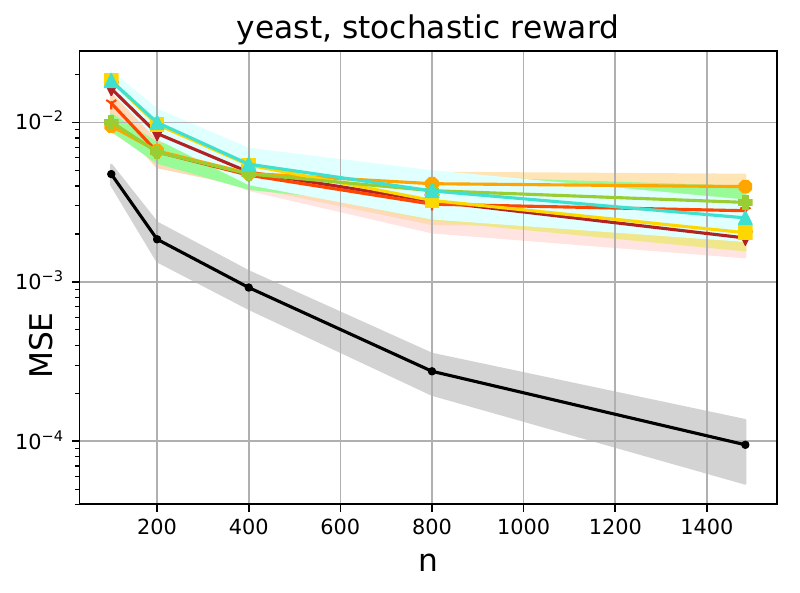}
	\end{minipage}
\end{figure*}

\begin{figure*}[ht]
    \centering
	\includegraphics[width=\linewidth]{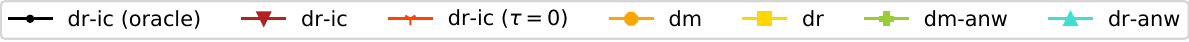}
	\centering
	\begin{minipage}{.24\textwidth}
		\centering
		\includegraphics[width=\linewidth]{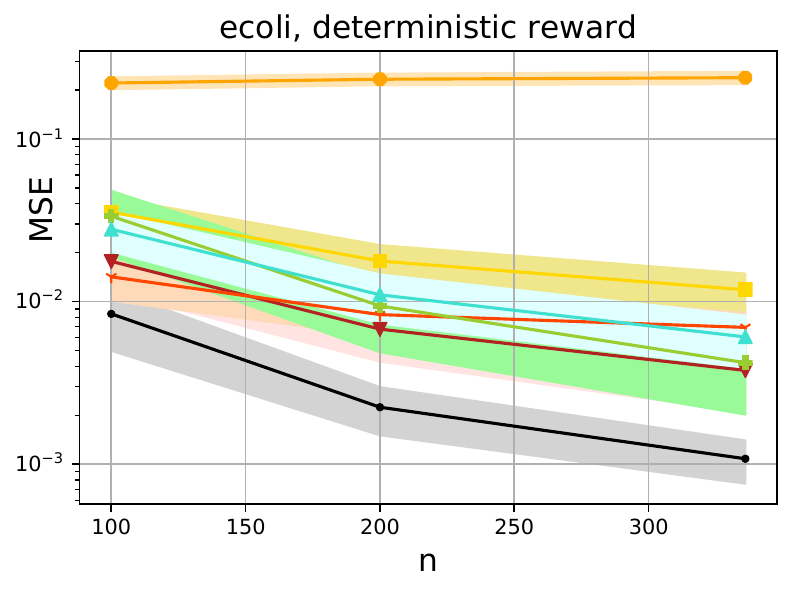}
	\end{minipage}
	\begin{minipage}{.24\textwidth}
		\centering
		\includegraphics[width=\linewidth]{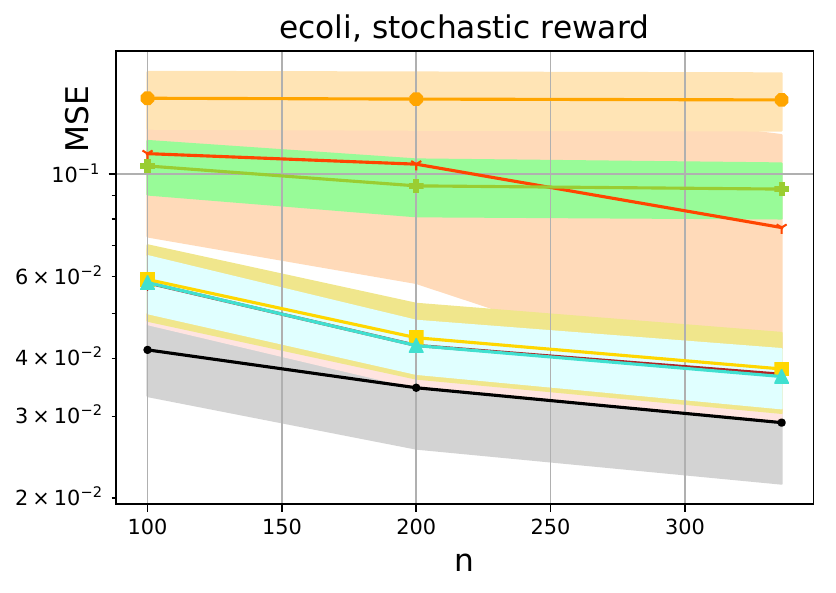}
	\end{minipage}
	\begin{minipage}{.24\textwidth}
		\centering
		\includegraphics[width=\linewidth]{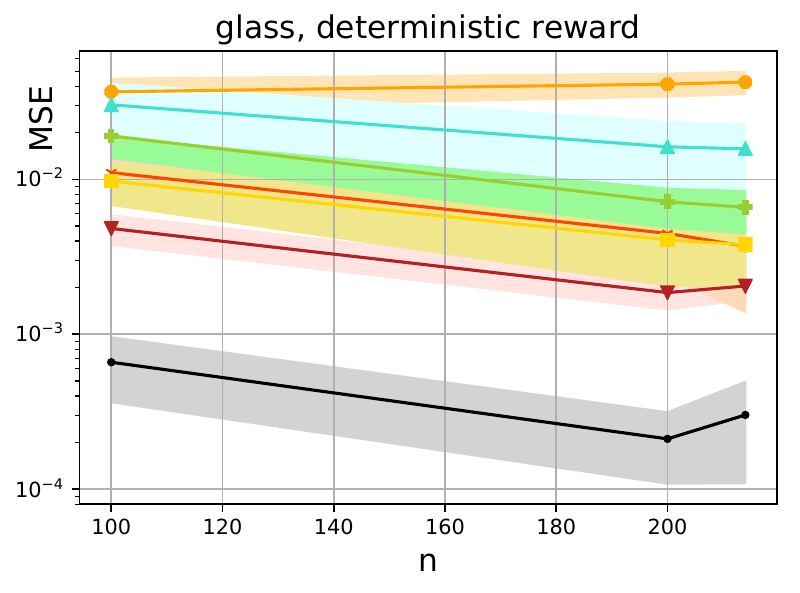}
	\end{minipage}
	\begin{minipage}{.24\textwidth}
		\centering
		\includegraphics[width=\linewidth]{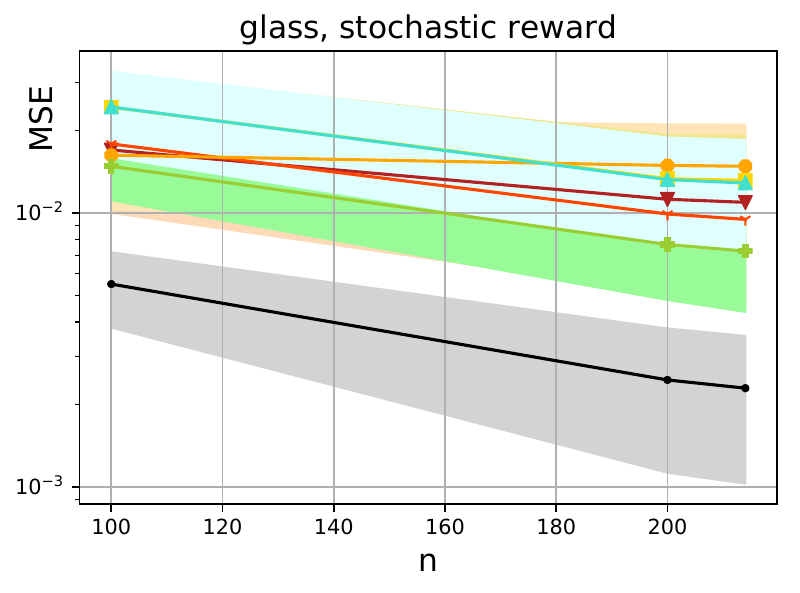}
	\end{minipage}
	\centering
	\begin{minipage}{.24\textwidth}
		\centering
		\includegraphics[width=\linewidth]{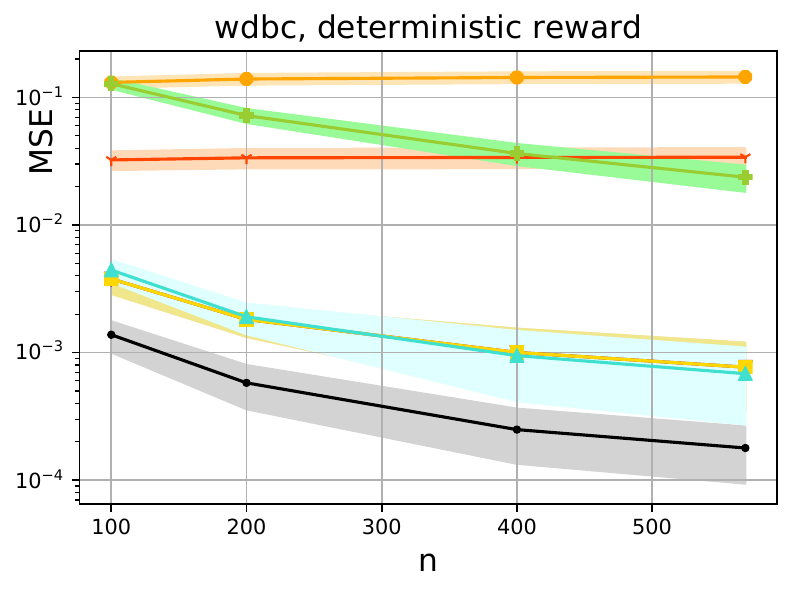}
	\end{minipage}
	\begin{minipage}{.24\textwidth}
		\centering
		\includegraphics[width=\linewidth]{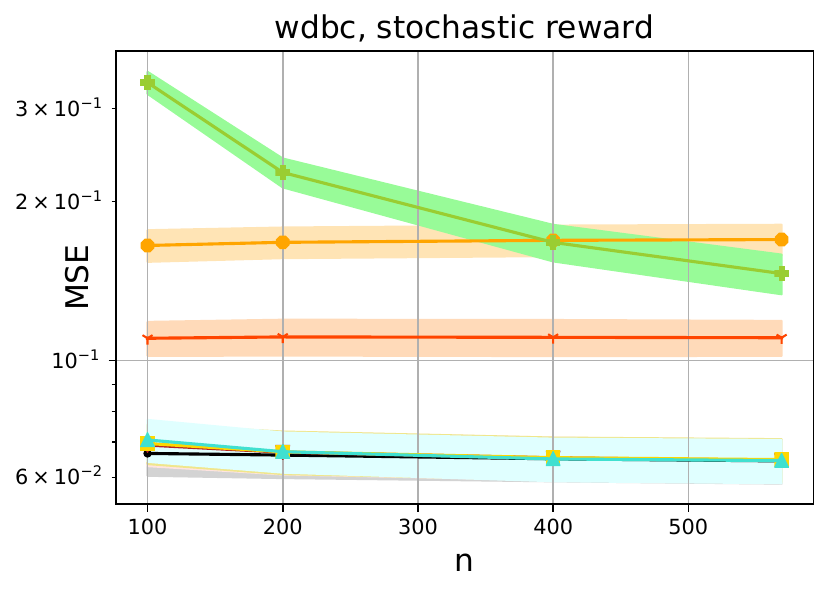}
	\end{minipage}
	\begin{minipage}{.24\textwidth}
		\centering
		\includegraphics[width=\linewidth]{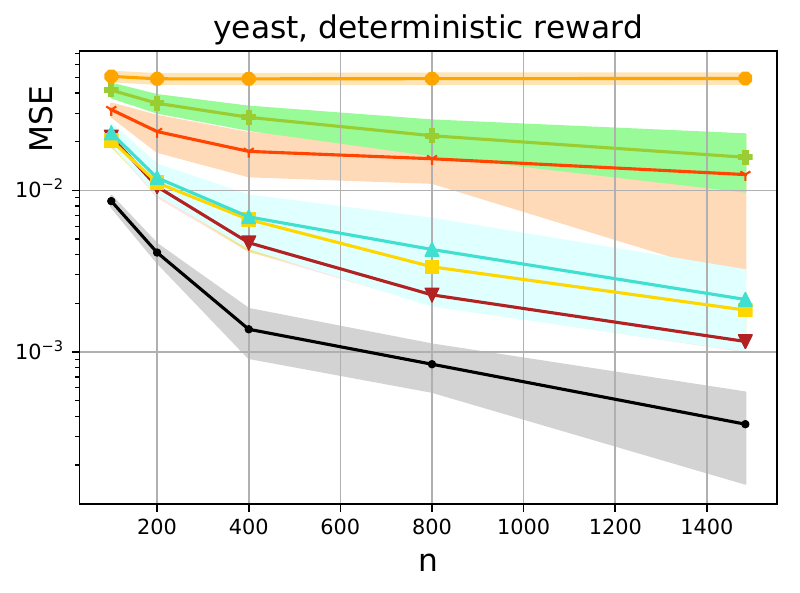}
	\end{minipage}
	\begin{minipage}{.24\textwidth}
		\centering
		\includegraphics[width=\linewidth]{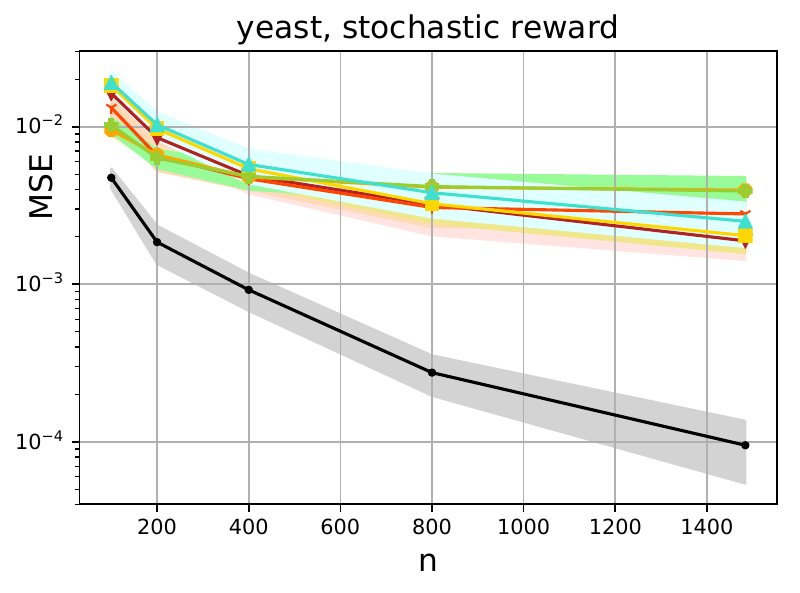}
	\end{minipage}
	\centering
	\begin{minipage}{.24\textwidth}
		\centering
		\includegraphics[width=\linewidth]{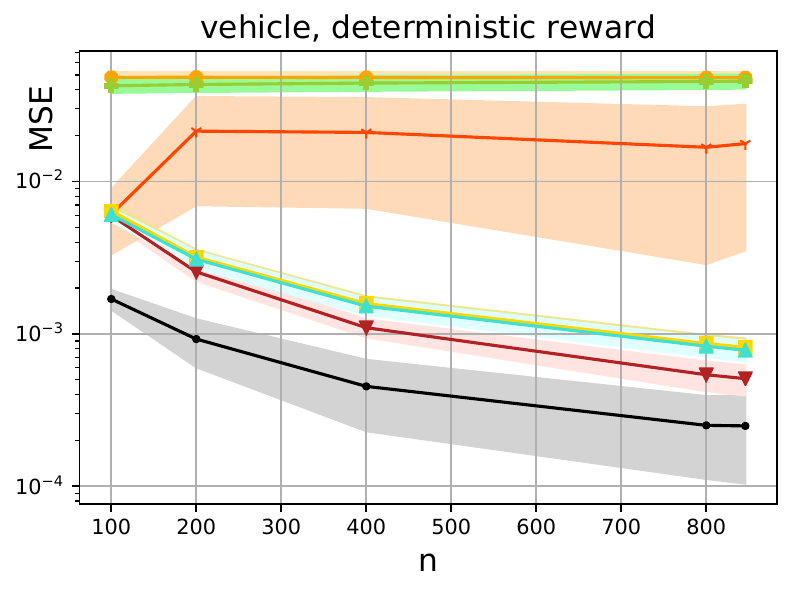}
	\end{minipage}
	\begin{minipage}{.24\textwidth}
		\centering
		\includegraphics[width=\linewidth]{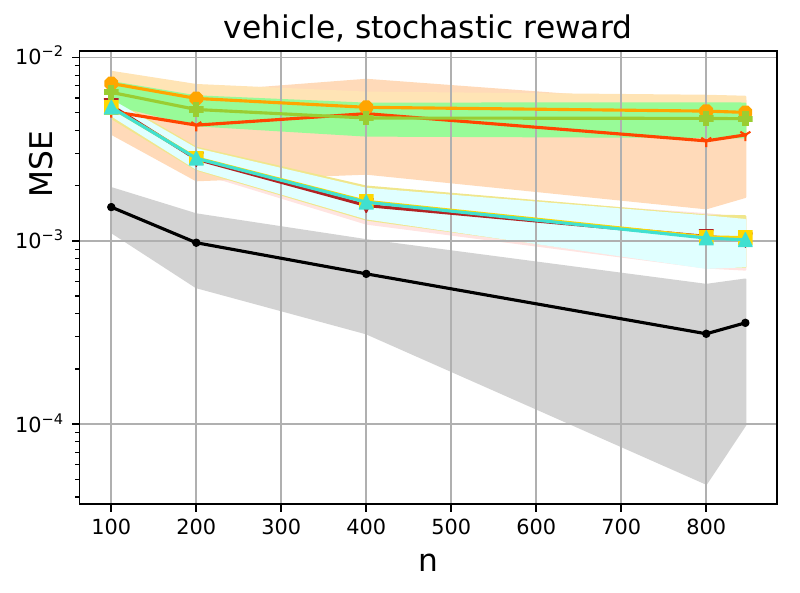}
	\end{minipage}
	\begin{minipage}{.24\textwidth}
		\centering
		\includegraphics[width=\linewidth]{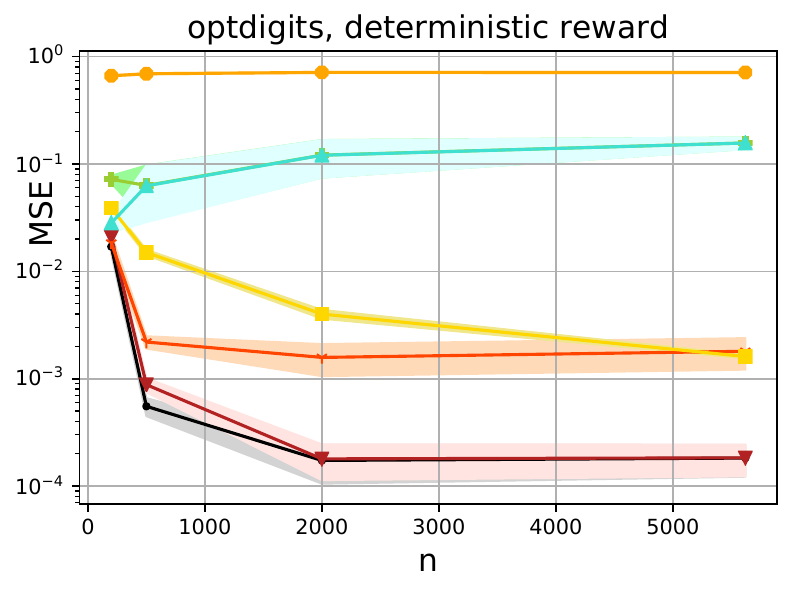}
	\end{minipage}
	\begin{minipage}{.24\textwidth}
		\centering
		\includegraphics[width=\linewidth]{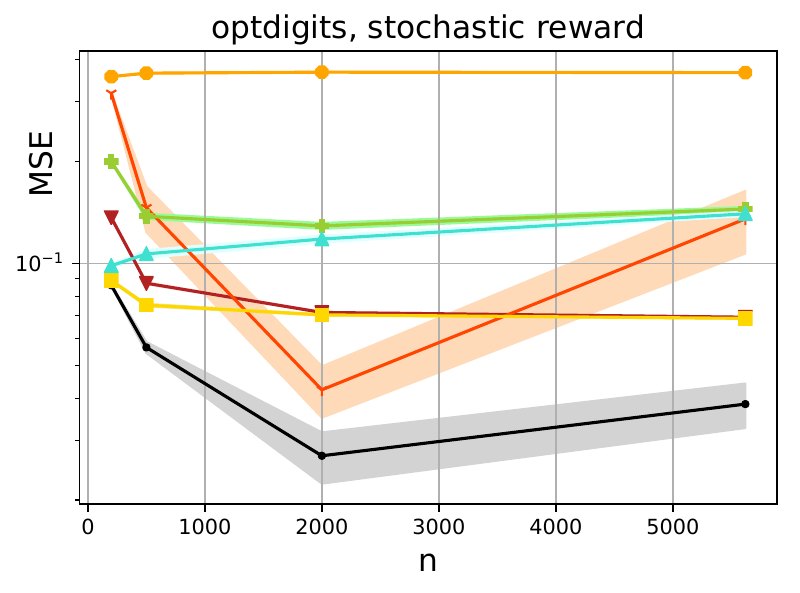}
	\end{minipage}
	\centering
	\begin{minipage}{.24\textwidth}
		\centering
		\includegraphics[width=\linewidth]{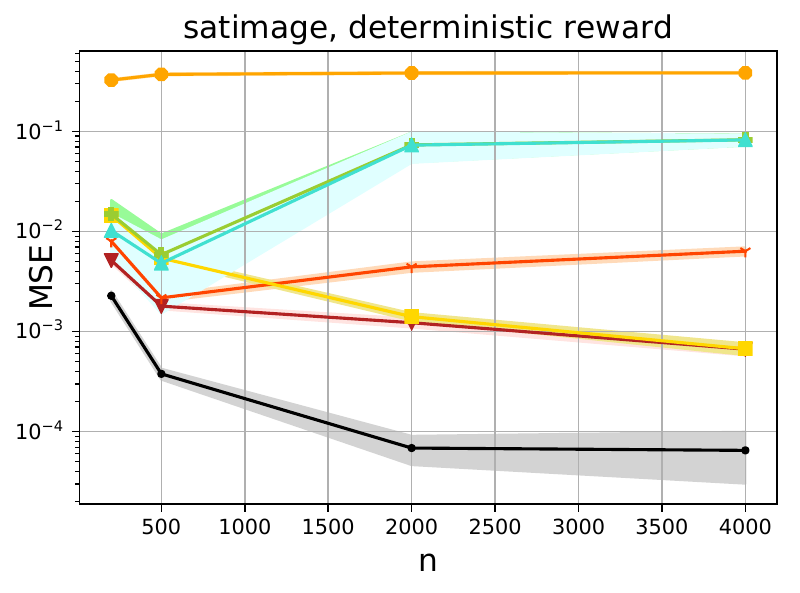}
	\end{minipage}
	\begin{minipage}{.24\textwidth}
		\centering
		\includegraphics[width=\linewidth]{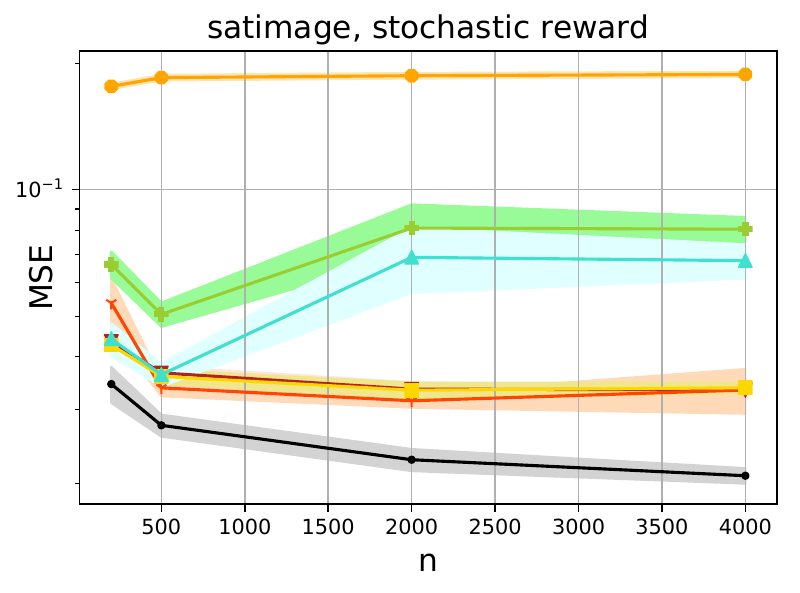}
	\end{minipage}
	\begin{minipage}{.24\textwidth}
		\centering
		\includegraphics[width=\linewidth]{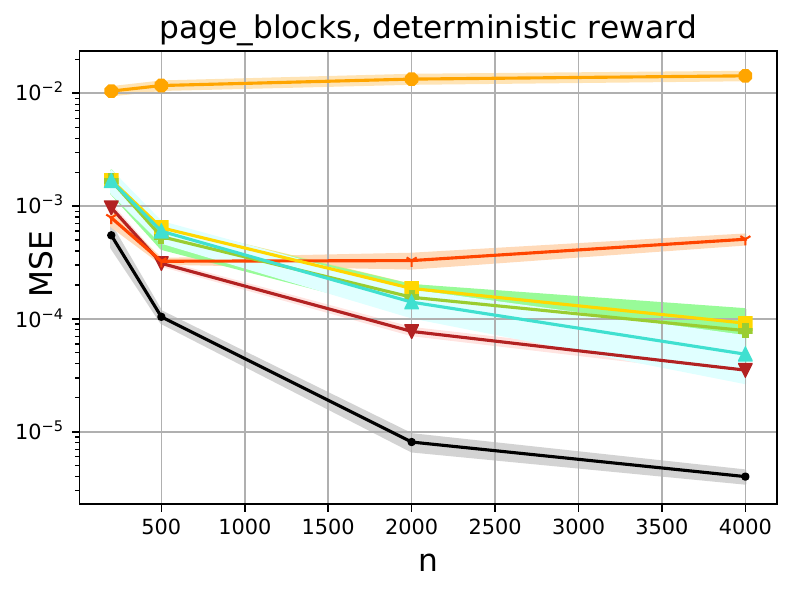}
	\end{minipage}
	\begin{minipage}{.24\textwidth}
		\centering
		\includegraphics[width=\linewidth]{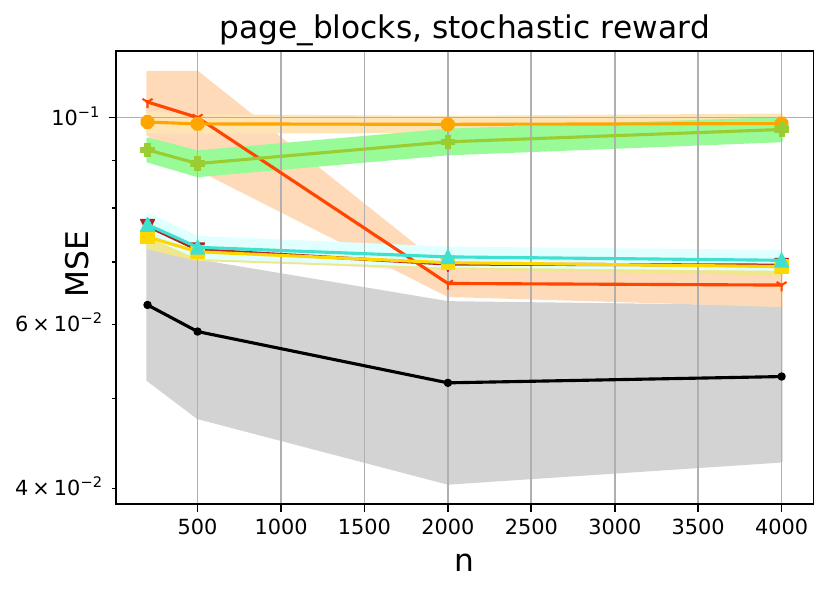}
	\end{minipage}
	\captionof{figure}{Adaptive NW based DM and DR comparisons with indicator approximation for action covariates.}
	\label{figcomp-nw-3}
\end{figure*}

We also  repeat the experiments in Fig.\ref{figcomp-nw-2} with an adaptive version of NW  estimator (labelled as dm-anw and dr-anw). That is, we use the adaptive radial basis function kernel $K((z_i - z_j)/h) = \exp(-h\| z_i-z_j \|_2/(w_i w_j))$ where the weights are as in Eq.\ref{covf}. The experiments still show superior performance of our approach in most cases, as illustrated in Fig.\ref{figcomp-nw-3}. However, we also note the improvement in NW estimate from the previous experiment. Thus, including the weights in the kernel leading to an adaptive nature improves the overall performance. The adaptive nature enables the algorithm to control the borrowing of information relative to the importance weights, which attributes to the improvement. Thus, irrespective of the kind of estimator we use, the effect of appropriately borrowing of information is the key to a better reward model, which can bring substantial improvement to practical performance of the off-policy evaluation.

%

\end{document}